\documentclass{article}

\usepackage{algorithm}
\usepackage{algpseudocode}

\usepackage{arxiv}

\usepackage{enumitem}
\usepackage{amsmath}
\usepackage{amssymb}
\usepackage{microtype}
\usepackage{graphicx}
\usepackage{booktabs} 
\usepackage{lmodern}
\usepackage[T1]{fontenc}
\usepackage{threeparttable}
\usepackage[table]{xcolor}

\usepackage{tikz}
\usetikzlibrary{arrows.meta, positioning, shapes, calc, decorations.pathmorphing, patterns}
\usetikzlibrary{backgrounds}

\usepackage{subcaption}
\usepackage{hyperref}
\usepackage{cleveref}
\usepackage{mathtools}
\usepackage{cuted} %
\usepackage{placeins} %

\usepackage{amsmath,amssymb,bm} %
\usepackage{boldline} %
\usepackage{pifont} %

\usepackage{hyperref} %

\usepackage{multirow}   
\usepackage[most]{tcolorbox}
\newtcolorbox{insightbox}[1][]{
  enhanced,
  colback=blue!3,
  colframe=blue!40!black,
  fonttitle=\bfseries\small,
  coltitle=blue!40!black,
  attach boxed title to top left={yshift=-2mm, xshift=3mm},
  boxed title style={colback=white, colframe=white},
  title={\faLightbulbO~Insight},
  left=4pt, right=4pt, top=4pt, bottom=4pt,
  boxrule=0.5pt,
  #1
}

\newtcolorbox{takeaway}[1][]{
  enhanced,
  colback=gray!5,
  colframe=gray!60!black,
  fonttitle=\bfseries\small,
  boxrule=0.6pt,
  left=5pt, right=5pt, top=5pt, bottom=5pt,
  arc=2pt,
  #1
}

\newtcolorbox{observationbox}[1][]{
  enhanced,
  colback=blue!4,
  colframe=blue!50!black,
  boxrule=0.5pt,
  left=5pt, right=5pt, top=4pt, bottom=4pt,
  arc=2pt,
  #1
}

\newtcolorbox{implicationbox}[1][]{
  enhanced,
  colback=green!4,
  colframe=green!50!black,
  boxrule=0.5pt,
  left=5pt, right=5pt, top=4pt, bottom=4pt,
  arc=2pt,
  #1
}

\newcounter{observation}
\newcounter{implication}

\newtcolorbox{observation}[1][]{
  enhanced,
  colback=blue!4,
  colframe=blue!50!black,
  boxrule=0.5pt,
  left=5pt, right=5pt, top=4pt, bottom=4pt,
  arc=2pt,
  before upper={\refstepcounter{observation}\textbf{O\theobservation:}~},
  #1
}

\newtcolorbox{implication}[1][]{
  enhanced,
  colback=green!4,
  colframe=green!50!black,
  boxrule=0.5pt,
  left=5pt, right=5pt, top=4pt, bottom=4pt,
  arc=2pt,
  before upper={\refstepcounter{implication}\textbf{I\theimplication:}~},
  #1
}

\newtcolorbox{promptbox}[2][]{
  enhanced,
  breakable,
  colback=gray!5,
  colframe=gray!70!black,
  fonttitle=\bfseries\small,
  title={#2},
  boxrule=0.6pt,
  left=5pt, right=5pt, top=5pt, bottom=5pt,
  arc=2pt,
  #1
}

\newtcolorbox{metricbox}[2][]{
  enhanced,
  colback=blue!3,
  colframe=blue!60!black,
  fonttitle=\bfseries\small,
  title={#2},
  boxrule=0.5pt,
  left=5pt, right=5pt, top=4pt, bottom=4pt,
  arc=2pt,
  #1
}

\usepackage{amsthm}

\usepackage{xcolor}

\begin{document}

\twocolumn[
\icmltitle{
Think Locally, Explain Globally: Graph-Guided LLM Investigations via Local Reasoning and Belief Propagation
}

\icmlsetsymbol{equal}{*}

\begin{icmlauthorlist}
\icmlauthor{Saurabh Jha}{ibm}
\icmlauthor{Rohan Arora}{ibm}
\icmlauthor{Bhavya}{ibm}
\icmlauthor{Noah Zheutlin}{ibm}
\icmlauthor{Paulina Toro Isaza}{ibm}
\icmlauthor{Laura Shwartz}{ibm}
\icmlauthor{Yu Deng}{ibm}
\icmlauthor{Daby Sow}{ibm}
\icmlauthor{Ruchi Mahindru}{ibm}
\icmlauthor{Ruchir Puri}{ibm}
\end{icmlauthorlist}

\icmlaffiliation{ibm}{IBM Research, Yorktown Heights, New York, USA}

\icmlcorrespondingauthor{Saurabh Jha}{Saurabh.Jha@ibm.com}

\icmlkeywords{Machine Learning, LLM Agents, Root Cause Analysis}

\vskip 0.3in

\begin{abstract}

LLM agents excel when environments are mostly static, and their required information fits in a model's context window, but they struggle with diagnostic investigations such as IT incident management, where operators iteratively mine massive observability data to identify the origins that explain observed symptoms, enabling correct remediation. The data in these domains is structured with inherent hidden dependencies: entities interact, signals co-vary, and the importance of a fact may only become clear after other evidence is discovered. To cope with bounded context windows, agents must summarize intermediate findings before their significance is known, increasing the risk of discarding key evidence. ReAct-style/CodeAct agents are especially brittle in this regime. Their retrieve-summarize-reason loop makes conclusions sensitive to exploration order and introduces run-to-run non-determinism, producing a reliability gap where Pass@k may be high, but Majority@k remains low. %
Increasing samples or reasoning lengths cannot guarantee stable results because, in high-risk diagnostic scenarios, consistent accuracy is required over sporadic success. Furthermore, the ReAct framework lacks a mechanism to revise its beliefs as new information becomes available.
In addition, ReAct entangles semantic reasoning with controller duties such as tool orchestration and state tracking; execution errors and plan drift degrade reasoning while consuming scarce context.
  We address these issues by formulating the investigation as abductive reasoning over a dependency graph and proposing EoG (Explanations over Graphs), an agentic architecture where an LLM performs bounded local evidence mining and labeling (cause vs symptom) while a deterministic controller manages traversal, state, and belief propagation to compute a minimal explanatory frontier. On representative ITBench diagnostics tasks, EoG improves accuracy and run-to-run consistency over ReAct baselines, including a 7\texttimes~ average gain in Majority@k F1 score.

\end{abstract}
]

\printAffiliationsAndNotice{} 

\section{Introduction}
\label{sec:intro}

LLM agents excel when environments are mostly static and their required information fits in the context window. They struggle with diagnostic investigations such as incident management, where operators iteratively mine massive observability data to identify the origins that explain observed symptoms, enabling correct remediation. 
This challenge affects numerous fields, including medical diagnosis~\cite{varatharajah2017eeggraph} and security threat analysis~\cite{wynn2014tara}. 
It is particularly critical in incident management, where cloud infrastructure and microservices constantly generate massive amounts of logs, metrics, traces, and events.
When failures occur, operators observe symptoms (alerts, errors, degradations) but must identify the minimal set of upstream sources whose effects propagate through the environment to explain observations; only then can they remediate. Since remediation based on incorrect diagnosis carries significant operational risk, the goal is to infer explanations from observational evidence rather than through intervention. More formally, this is \emph{abductive reasoning} over massive observability data from an environment with inherent graphical structure composed of entities connected by dependency edges~(\S\ref{sec:problem-statement}).

\textbf{Limits of existing agent patterns:}
Current agentic approaches struggle with abductive reasoning. For example, on the SRE scenarios of ITBench~\cite{jhaitbench}\footnote{ITBench is an open-source benchmark for testing how well AI agents can diagnose and resolve IT issues.} (pertaining to incident management), the best ReAct agent achieves only 13.81\% pass@1 recall (\S\ref{sec:motivation}).  Beyond well-known limitations such as tool invocation failures and redundant or degenerate tool calls ~\cite{cemri2025multiagentllmsystemsfail}, ReAct~\cite{yao2023react} and CodeAct~\cite{wang2024codeact}-based agents struggle significantly with context gathering and action sequencing under uncertainty. Reasoning is only as sound as the evidence it conditions on: if relevant context is not retrieved, it cannot be reasoned over. In open-ended investigations, evidence relevance is not known \emph{a priori}, and ReAct agents lack mechanisms for systematic context acquisition. Tool calls are issued opportunistically, missing low-frequency, high-impact signals or retrieving excessive irrelevant data. Summarizing the retrieved evidence before reasoning introduces a critical failure mode that discards information before relevance can be established, often losing critical evidence ~\cite{liu2024lost}. 

Furthermore, open-ended investigations exhibit non-monotonic belief revision behaviors: interpretation of entities with the related signals naturally changes as new evidence emerges elsewhere in the system, making reasoning order-dependent. The sequence of evidence retrieval, summarization, and presentation to the LLM materially affects conclusions. In practice, ReAct agents show high variance across runs due to planning non-determinism, unstable exploration orderings, and lossy summarization, producing fragile, incomplete, irreproducible explanations. These failures are architectural: ReAct implicitly assumes linear, monotonic reasoning over an accumulating context, whereas abductive reasoning over massive graphs requires structured context acquisition with explicit belief revision.

\textbf{Towards graph-guided agentic explanations:}
We introduce an operational-graph abstraction for investigations under partial observability.  We represent the environment as an operational graph and define diagnosis as \emph{explanations over such dynamically evolving operational graphs}: explanations correspond to minimal subgraphs consistent with observed evidence. Framing diagnosis in this way converts an ill-defined exploration problem into structured inference. For example, across IT domains, abductive explanations are naturally structured by dependency relationships among system entities.

"Diagnosis is non-monotonic because new evidence can change what we believe about an entity's role. Drawing from belief propagation~\cite{pearl1988probabilistic}, we model diagnosis as an iterative local belief assignment, revision, and propagation process through the graph. This leads to a disaggregated architecture: local reasoning occurs via LLMs over bounded neighborhoods, while a deterministic controller manages traversal order, context gathering, belief bookkeeping, and termination. This separation enables non-monotonic belief revision through explicit recording and updating of beliefs rather than implicit overwriting in growing ReAct-style prompts. By externalizing planning and context management, the system keeps context bounded, reduces tool-use failures, and produces reproducible, auditable explanations. This separation also addresses production constraints: bounded per-node context scales to large datasets, while deterministic control minimizes LLM calls and API rate limit exposure. Our contributions:
\begin{itemize}[leftmargin=*,nosep]
    \item \textbf{EoG agent with separation of concerns:} We develop EoG (Explanations over Graphs), a disaggregated architecture where a \emph{Deterministic Controller} manages graph traversal, state, and tool orchestration, while a stateless \emph{Abductive Policy} $\pi_{\text{abd}}$ classifies each node as origin, symptom, or healthy based on bounded local evidence (\S\ref{sec:method}). This separation eliminates controller failures, ensures auditability, and enables systematic exploration.

    \item \textbf{Semantic Belief Propagation:} We introduce SBP, a message-passing mechanism inspired by belief propagation~\cite{pearl1988probabilistic}, to address non-monotonic belief revision. When new evidence changes a node's belief, it broadcasts to neighbors, triggering reconsideration by $\pi_{\text{abd}}$---correcting early misdiagnoses without restart~(\S\ref{sec:algorithm}).

    \item \textbf{Validation:} On ITBench SRE scenarios~\cite{jhaitbench}, EoG achieves 7$\times$ higher Majority@$k$ F1 than ReAct baselines (\S\ref{sec:results}). We have extended this formalism to SWE tasks to diagnose production bugs~\cite{cui2025agentic}.
\end{itemize}

\section{Problem Formulation}
\label{sec:problem}
\label{sec:problem-statement}

We model the environment as a \emph{Dependency Graph} $G=(V,E)$ where nodes $V$ are \emph{entities} and edges $E$ encode structural dependencies along which effects propagate. Entity examples include IT services in SRE settings, hosts in security, or physiological signals in medical diagnosis. Entities can originate faults, exhibit symptoms, or propagate effects along domain dependency edges.
In this work, we focus on abductive reasoning in SRE to diagnose failures from observed anomalies, where $V$ are services, pods, and configurations, and $E$ are API calls or ownership relations.

An \emph{anomaly} is an observation of abnormal behavior at an entity; in SRE, \emph{alerts} are anomalies that operators configure to receive notifications.
Given anomalies $\mathcal{O}$ on entities $V_\mathcal{O} \subseteq V$, the task is to \emph{explain them}. %
Formally, we seek:
\begin{enumerate}[nosep]
\item An \emph{Explanatory Subgraph} $G_S = (V_S, E_S)$: $V_S \subseteq V$ is the active set; $E_S$ are directed edges representing inferred propagation: $u \to v$ means ``$u$ explains $v$''.
\item A \emph{Frontier} $\mathcal{F} \subseteq V_S$: the minimal origin set whose faults explain all anomalies on $V_\mathcal{O}$.
\end{enumerate}
Our focus is explanatory inference from observational evidence. We avoid interventions since remediation actions carry operational risk. 
This formulation naturally extends to other domains such as FinOps, DevOps, healthcare, and forensics.

\section{Characterizing Failure Modes}
\label{sec:motivation}
We motivate our approach by characterizing failure modes of ReAct agents~\cite{yao2023react} built using Codex~\cite{ wang2024codeact, openai_codex} (Appendix~\ref{appendix:react_agent}) on the SRE scenarios of ITBench~\cite{jhaitbench}.%
We focus on SRE scenarios as they require abductive reasoning---inferring root causes from observed symptoms.

\subsection{The Reliability Gap}
\label{sec:motivation-variance}

\begin{figure}[t]
    \centering
    \includegraphics[width=0.8\linewidth]{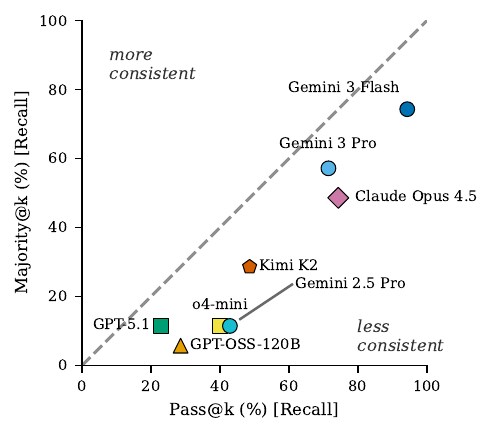}
    \caption{The reliability gap between Pass@$k$ (success in at least one run) and Majority@$k$ (success in the majority of runs). Significant deviations from dotted line indicate stochastic success rather than consistent reasoning.}
    \label{fig:pass_vs_majority}
    \vspace{-0.6cm}
\end{figure}
While most models perform poorly under the ReAct paradigm, we quantify a more fundamental issue: stochastic success rather than consistent reasoning.
We define the \emph{reliability gap} as Pass@$k$ (success in at least one of $k$ runs) minus Majority@$k$ (success in the majority of runs).\footnote{When verification is cheap (e.g., code generation with test suites), one can sample until a correct answer is verified, making Pass@$k$ sufficient for accuracy. However, even then, low Majority@$k$ has consequences for maintainability, latency, and predictability. In incident management, verification requires remediation actions that carry operational risk, so consistent correctness is essential.}
Figure~\ref{fig:pass_vs_majority} shows this gap ranging from 14 points for Gemini~3~Flash to 32 points for Gemini~2.5~Pro.
Even the best model (Gemini~3~Flash: 89\% Pass@$k$, 74\% Majority@$k$) exhibits significant variance, with only 51\% of scenarios yielding consistent correct diagnoses across all runs.
This reliability gap must be addressed to ensure successful deployment of agents in enterprise settings~\cite{kartik2025agentcompassreliableevaluationagentic}.

\subsection{Exploration Failures}
\label{sec:motivation-exploration}

\begin{figure*}[t]
  \centering
  \begin{subfigure}[t]{0.32\textwidth}
    \centering
    \vspace{0pt}
    \includegraphics[width=\linewidth]{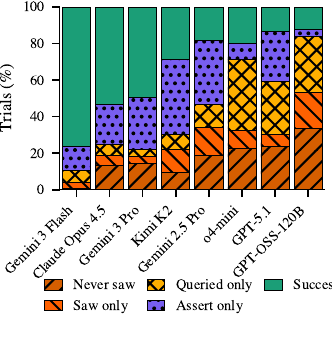}
    \caption{GT discovery funnel.}
    \label{fig:conversion_funnel}
  \end{subfigure}
  \hfill
  \begin{subfigure}[t]{0.32\textwidth}
    \centering
    \vspace{0pt}
    \includegraphics[width=\linewidth]{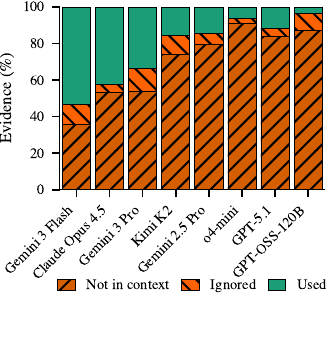}
    \caption{Evidence discovery failures.}
    \label{fig:failure_modes}
  \end{subfigure}
  \hfill
  \begin{subfigure}[t]{0.32\textwidth}
    \centering
    \vspace{0pt}
    \includegraphics[width=\linewidth]{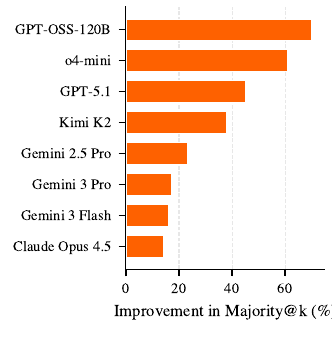}
    \caption{Oracle reordering impact.}
    \label{fig:oracle_reordering}
  \end{subfigure}
  \caption{ReAct agent failure modes. (a)~GT discovery funnel: \emph{never saw} (exploration), \emph{saw not queried} (attention), \emph{queried not asserted} (reasoning), \emph{asserted not in output} (prioritization), \emph{asserted in output} (success). (b)~Evidence discovery: \emph{not in context} (red), \emph{ignored} (orange), \emph{used} (green). (c)~Impact of oracle reordering on Majority@$k$.}
  \label{fig:exploration_pathology}
\end{figure*}

Abductive reasoning in complex systems is reduced to a search over a dependency graph of entities.
Given observed symptoms, the goal is to identify the minimal upstream entities whose states explain all observations.
This search space is exponential; the dependency graph constrains tractable exploration.
Our analysis reveals that ReAct agents fail not because they cannot reason, but because they lack structural priors to use a graph.

\textbf{Ground truth (GT) entity discovery failures.}
Figure~\ref{fig:conversion_funnel} tracks the GT entity through a discovery funnel.
A substantial failure mode is ``GT never saw'': the GT entity is linked to the symptom via dependency edges, yet the ReAct agent fails to follow these links.

\textbf{Evidence discovery failures.}
Figure~\ref{fig:failure_modes} shows a high incidence of \emph{Evidence not in context}: supporting data existed but was never fetched, either because the entity was never explored or because what constitutes ``relevant'' evidence depends on information discovered later (\S\ref{sec:motivation-ordering}).
The small ``evidence ignored'' category indicates that modern LLMs can use evidence when it is present; the bottleneck is discovery, not reasoning.

\textbf{Path dependence.}
\label{sec:motivation-ordering}
ReAct agents are highly sensitive to information ordering.
In an oracle experiment (Figure~\ref{fig:oracle_reordering}), reordering investigation sequences to prioritize paths from symptoms to root causes without adding new information, yields 14--70\% improvement in Majority@$k$ across LLMs.
This path dependence indicates agents struggle with \emph{non-monotonic reasoning}: early wrong beliefs persist despite emerging evidence.

\subsection{Controller Failures}
\label{sec:motivation-controller}

ReAct agents experience frequent controller failures---plan abandonment, tool repetition, and syntactic errors.
We use MAST~\cite{mast2025} to characterize these failures on our agent trajectories (Appendix~\ref{appendix:controller_analysis}).
The key implication is the separation of deterministic control flow (tool orchestration, state management) from non-deterministic reasoning (causal inference).

These failure modes reinforce one another.
Controller brittleness consumes context that could otherwise support broader exploration.
Incomplete exploration leads to missed evidence.
Path-dependent reasoning prevents recovery when initial assumptions prove incorrect.

\section{Methodology}
\label{sec:method}
\begin{figure}[!t]
    \centering
    \begin{tikzpicture}[
        node distance=0.5cm,
        every node/.style={font=\footnotesize},
        box/.style={draw, rounded corners=2pt, minimum height=0.45cm, font=\scriptsize, align=center},
        data/.style={box, fill=gray!10},
        llm/.style={box, fill=orange!15, draw=orange!60!black},
        ctrl/.style={box, fill=green!10, draw=green!60!black},
        gnode/.style={circle, draw, minimum size=0.3cm, inner sep=0pt},
        origin/.style={gnode, fill=red!25, draw=red!60!black, double, double distance=0.8pt, line width=0.5pt},
        symptom/.style={gnode, fill=orange!25, draw=orange!60!black, line width=0.6pt, 
            postaction={pattern=north east lines, pattern color=orange!60!black}},
        ghost/.style={gnode, draw=black!50, text=black!70},
        arr/.style={-{Stealth[length=1.2mm]}, semithick},
        discovered/.style={-{Stealth[length=1.2mm]}, semithick, draw=green!60!black, densely dashed},
        msg/.style={-{Stealth[length=1mm]}, draw=blue!50, semithick, decorate, decoration={snake, amplitude=0.3mm, segment length=1.2mm}},
        phase/.style={font=\scriptsize\bfseries, text=black!80},
        note/.style={font=\scriptsize, text=black!60, align=center},
    ]

    \node[phase, anchor=west] at (-3.5, 0.3) {1. Topology $G$};
    
    \node[ghost, label={[font=\scriptsize]above:$S_1$}] (s1t) at (-3.4, -0.5) {};
    \node[ghost, label={[font=\scriptsize]above:$S_2$}] (s2t) at (-2.6, -0.5) {};
    \node[ghost, label={[font=\scriptsize]above:$S_3$}] (s3t) at (-1.4, -0.5) {};
    \node[ghost, label={[font=\scriptsize]above:$S_4$}] (s4t) at (-0.6, -0.5) {};
    
    \node[ghost, draw=black!30, label={[font=\scriptsize]below:$S_5$}] (h1) at (-3.4, -1.2) {};
    \node[ghost, draw=black!30, label={[font=\scriptsize]below:$S_6$}] (h2) at (-2.6, -1.2) {};
    \node[ghost, draw=black!30, label={[font=\scriptsize]below:$S_7$}] (h3) at (-0.6, -1.2) {};
    
    \draw[black!50, semithick] (s1t)--(s2t);
    \draw[black!50, semithick] (s3t)--(s4t);
    \draw[black!30] (s1t)--(h1);
    \draw[black!30] (s2t)--(h2);
    \draw[black!30] (s4t)--(h3);
    \draw[black!30] (h1)--(h2);
    
    \node[note, text=red!60!black] at (-2.0, -0.85) {$S_2$--$S_3$\\missing};
    \node[data, fill=red!10, draw=red!40, font=\scriptsize] at (-2.0, -2.05) {Alert on $S_2$};
    
    \node[phase, anchor=west] at (-3.5, -3.1) {3. Explanatory Graph};
    
    \node[origin, label={[font=\scriptsize]above:$S_1$}] (s1f) at (-3.4, -3.75) {};
    \node[symptom, label={[font=\scriptsize]above:$S_2$}] (s2f) at (-2.6, -3.75) {};
    \node[symptom, label={[font=\scriptsize]above:$S_3$}] (s3f) at (-1.4, -3.75) {};
    \node[symptom, label={[font=\scriptsize]above:$S_4$}] (s4f) at (-0.6, -3.75) {};
    
    \draw[arr, draw=black!70] (s2f) -- (s1f);
    \draw[discovered] (s3f) -- (s2f);
    \node[font=\scriptsize, text=green!50!black] at (-2.0, -3.4) {new};
    \draw[arr, draw=black!70] (s4f) -- (s3f);
    \draw[msg] (s4f.south) to[bend right=25] (s3f.south);
    
    \node[note, anchor=west] at (-3.5, -4.4) {$\mathcal{F} = \{S_1\}$ (flash sale)};
    
    \node[phase] at (1.8, 0.3) {2. EoG};
    
    \node[data, fill=red!10, draw=red!40, font=\scriptsize, minimum width=1.1cm] (active) at (1.8, -0.4) {ActiveSet};
    \node[ctrl, font=\scriptsize] (pop) at (1.8, -1.3) {Pop $v$};
    \node[ctrl, font=\scriptsize, minimum width=0.9cm] (cxc) at (0.9, -2.1) {CxC};
    \node[llm, font=\scriptsize, minimum width=0.9cm] (policy) at (2.7, -2.1) {$\pi_{\text{abd}}$};
    \node[ctrl, font=\scriptsize, minimum width=1.1cm] (update) at (1.8, -2.9) {Update $G_S$};
    \node[data, font=\scriptsize] (expand) at (0.9, -3.7) {Expand};
    \node[data, font=\scriptsize] (prop) at (2.7, -3.7) {Propagate};
    
    \draw[arr] (active) -- (pop);
    \draw[arr] (pop) -- (cxc);
    \draw[arr] (cxc) -- (policy);
    \draw[arr] (policy) -- (update);
    \draw[arr] (update) -| (expand);
    \draw[arr] (update) -| (prop);
    \draw[arr, dashed] (expand.west) -- ++(-0.15,0) |- (active.west);
    \draw[arr, dashed] (prop.east) -- ++(0.15,0) |- (active.east);
    
    \begin{scope}[yshift=-5.0cm, xshift=-0.3cm]
        \node[origin, minimum size=0.25cm] at (-2.8, 0) {};
        \node[right, font=\scriptsize] at (-2.6, 0) {Origin};
        \node[symptom, minimum size=0.25cm] at (-1.2, 0) {};
        \node[right, font=\scriptsize] at (-1.0, 0) {Symptom};
        \draw[discovered] (0.4, 0) -- (0.8, 0);
        \node[right, font=\scriptsize] at (0.9, 0) {Discovery};
        \draw[msg] (2.5, 0) -- (2.9, 0);
        \node[right, font=\scriptsize] at (3.0, 0) {Msg};
    \end{scope}
    
    \end{tikzpicture}
    \vspace{-2mm}
    \caption{EoG agent architecture. \textbf{(1)} Topology $G$ with missing edge; alert seeds ActiveSet. \textbf{(2)} Loop: pop node, fetch context (CxC), run policy $\pi_{\text{abd}}$, update $G_S$. \emph{Expand} adds new neighbors; \emph{Propagate} reactivates on belief change. \textbf{(3)} Output $G_S$ with discovered edge and frontier $\mathcal{F}=\{S_1\}$.}
    \label{fig:eog-architecture}
    \end{figure}
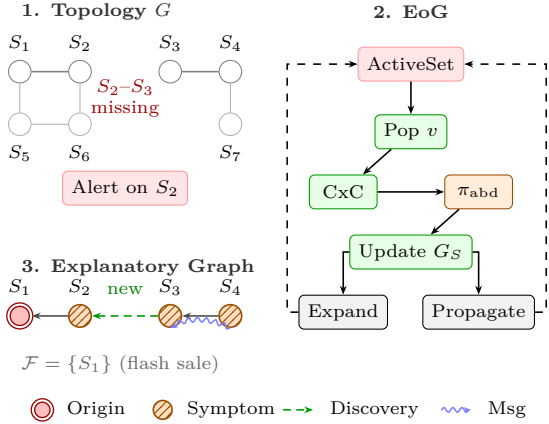
    \vspace{-2mm}
    
We present \textbf{EoG (Explanations over Graphs)}, an agent that traverses the dependency graph $G$ to identify the \emph{frontier} $\mathcal{F}$: the minimal set of origin nodes whose faults explain all observed anomalies (\S\ref{sec:problem-statement}). EoG builds an \emph{Explanatory Graph} $G_S$ where each visited node carries a \emph{belief}: its inferred explanatory role (\textsc{Origin}, \textsc{Symptom}, or \textsc{Healthy}) and supporting evidence. The agent addresses the limitations of ReAct (\S\ref{sec:motivation}) through two design principles:

\begin{enumerate}[nosep]
    \item \textbf{Separation of decision and control.} A \emph{Deterministic Controller} handles graph traversal, state management, and tool orchestration, while a stateless \emph{Abductive Policy} $\pi_{\text{abd}}$ performs local inference over raw evidence (e.g., logs, metrics retrieved via tools) and neighbor messages for each node $v \in V_S$. This separation eliminates controller failures (\S\ref{sec:motivation-controller}), ensures auditability, and enforces systematic graph exploration.
    
    \item \textbf{Belief revision via Semantic Belief Propagation (SBP).} Initial beliefs may be incorrect due to incomplete information. For instance, a node with strong failure symptoms (e.g., out-of-memory) may be a victim of upstream problems (e.g., increased load). When new evidence changes a node's belief, it broadcasts to its neighbors, triggering re-evaluation. This message-passing mechanism, inspired by belief propagation~\cite{pearl1988probabilistic}, addresses path dependence (\S\ref{sec:motivation-ordering}), by correcting  early misdiagnoses as evidence propagates, without restarting the investigation.
\end{enumerate}

\subsection{Architecture Overview}

The architecture comprises three modules (Figure~\ref{fig:eog-architecture}):

\textbf{The Deterministic Controller.}
A symbolic engine that maintains all persistent state: the \textbf{Explanatory Graph} $G_S = (V_S, E_S)$ with node beliefs, the \textbf{Global Ledger} $\mathcal{L}$ (immutable history of reasoning steps), the \textbf{ActiveSet} (nodes pending investigation), and \textbf{message queues}, which 
execute graph traversal, message routing, and termination logic without semantic reasoning. This separation ensures \textbf{safety and reproducibility}: the agent cannot execute arbitrary code; it can only request data via the Context Contract (a specialized tool described below). It also provides \textbf{operational efficiency}: all bookkeeping requires no LLM calls, bounding inference costs to policy invocations. 

\textbf{The Context Contract (CxC).}
A strict interface that bounds and focuses the information provided to the policy, drawing on principles of active context management~\cite{li2025sculptor, zhang2025agentic}. The CxC is implemented as an MCP tool~\cite{anthropic2024mcp} with a simple interface \texttt{get\_context(entity\_id, start\_time, end\_time)}, abstracting away domain-specific details. When investigating node $v$, the CxC returns local evidence on that entity (e.g., logs, metrics, traces in IT systems) and topological context (immediate neighbors). The Controller bundles this with the Inbox to form the \textbf{Context Packet}.

\emph{Relevance Filtering:} Domain experts can equip the CxC with specialized tools (e.g., ``fetch top-k log anomalies'') to minimize evidence omission. This ensures the policy sees the most relevant signal without context overflow.

\textbf{The Abductive Policy ($\pi_{\text{abd}}$).}
A stateless, LLM-parameterized policy that performs local abductive inference. Given a bounded Context Packet for a node $v$, $\pi_{\text{abd}}$ produces:
(i) a local belief $B_v$,
(ii) propagation claims $\mathcal{P}_v$, and
(iii) a set of candidate entities $\mathcal{N}_{next}$ for further investigation.
While the Controller governs execution and termination, $\pi_{\text{abd}}$ determines the epistemic trajectory of the search. We use ``policy'' in the decision-theoretic sense: a mapping from local observations to belief assignment.

\subsection{State Formalism}
\label{sec:formalism}

EoG maintains following state variables:

\textbf{1. The Global Ledger} ($\mathcal{L}$): an append-only log recording each $(v, B_v, t)$ tuple tracking which node was evaluated, its resulting belief, and when. This scratchpad~\cite{nye2021show} enables replay, debugging, and auditability. 

\textbf{2. Local Belief ($B_v$)}: stored per node in $G_S$, updated after each policy execution.
$B_v = (L_v, \mathcal{E}_v)$ where:
\begin{itemize}[nosep]
    \item $L_v \in \{\textsc{Healthy}, \textsc{Origin}, \textsc{Symptom}, \textsc{Defer}\}$ is the explanatory label:
    \begin{itemize}[nosep]
        \item \textsc{Healthy}: Node is operating normally within the incident window.
        \item \textsc{Origin}: Node has a citable change (config, deployment, resource) that explains the incident.
        \item \textsc{Symptom}: Node is degraded, but evidence points to an upstream cause.
        \item \textsc{Defer}: Evidence is inconclusive---either absent or unrecognized by the policy.
    \end{itemize}
    \item $\mathcal{E}_v$ is the evidence summary.
\end{itemize}

\textbf{3. Messages ($m_{u \to v}$)}: stored in per-node Inboxes. A message contains the sender's belief $B_u$ at the time of broadcast. When node $v$ is evaluated, the policy reads its Inbox to reason about neighbor states, exploiting the Local Markov Property~\cite{pearl1988probabilistic}.

\subsection{Algorithm}
\label{sec:algorithm}

We present the algorithmic steps of the EoG agent, which interleaves exploration (discovery of new nodes) and propagation ( belief refinement). 

\subsubsection{Initialization (Bootstrap)}
The Controller initializes the investigation:
\begin{itemize}[nosep]
    \item Define time window $W$ (e.g., incident start/end time).
    \item Initialize:
    \begin{itemize}[nosep]
    \item $\text{ActiveSet} \leftarrow V_\mathcal{O}$ (e.g., alerting entities)
    \item $V_S \leftarrow \emptyset, E_S \leftarrow \emptyset$
    \item $\text{Budget} \leftarrow \text{max investigation hops}$
    \end{itemize}
\end{itemize}

\subsubsection{Active Inference Loop}
The core loop processes nodes in the ActiveSet until convergence (no new messages) or budget exhaustion (tracking the maximum number of investigation hops).

\begin{algorithmic}[1]
\While{$\text{ActiveSet} \neq \emptyset \land \text{Budget} > 0$}
    \State $v \gets \text{ActiveSet}.\textsc{Pop}()$
    \State $V_S.\text{add}(v)$
    
    \State \textbf{// 1. Fetch Context}
    \State $\text{Inbox} \gets \text{CollectMessages}(v)$
    \State $\text{LocalContext} \gets \text{CxC.Fetch}(v, W)$
    \State $\text{Packet} \gets (\text{LocalContext}, \text{Inbox})$
    
    \State \textbf{// 2. Execute Policy}
    \State $(B_v^{new}, \mathcal{P}_v, \mathcal{N}_{next}) \gets \pi_{\text{abd}}(\text{Packet})$
    
    \State \textbf{// 3. Damping (Heuristic)}
    \If{$\textsc{Flips}(v) > k_{\text{thresh}}$}
        \State $B_v^{new}.L \gets \textsc{Defer}$
    \EndIf
    
    \State \textbf{// 4. Update State}
    \State $\text{Ledger}.\text{append}(v, B_v^{new})$
    \For{$(u \to v) \in \mathcal{P}_v$}
        \State $E_S.\text{add}(u \to v)$
        \If{$u \notin V_S$} \State $\text{ActiveSet}.\text{add}(u)$ \Comment{Exploration} \EndIf
    \EndFor
    
    \State \textbf{// 5. Propagation}
    \If{$B_v^{new} \neq B_v^{old}$}
        \State $\text{Broadcast}(B_v^{new} \text{ to } \mathcal{N}(v))$
        \State $\text{ActiveSet}.\text{add}(\mathcal{N}(v))$ \Comment{Re-activate neighbors}
    \EndIf
    
    \State \textbf{// 6. Propose (Optional)}
    \For{$u \in \mathcal{N}_{next}$}
        \If{$u \notin V_S$} \State $\text{ActiveSet}.\text{add}(u)$ \EndIf
    \EndFor

    \State $\text{Budget}--$ \Comment{Decrement Budget}
\EndWhile
\State \textbf{// Post-processing: Compute Frontier}
\State $\mathcal{F} \gets \{v \in V_S : L_v = \textsc{Origin} \land \nexists u \in V_S, L_u = \textsc{Origin} \land u \rightsquigarrow v\}$
\State \textbf{// Finalize: Generate Report}
\State $\text{Report} \gets \pi_{\text{finalize}}(G_S, \mathcal{F}, \mathcal{L})$
\end{algorithmic}

The loop terminates when no more messages are pending (fixed point) or the budget is exhausted.
The Frontier $\mathcal{F}$ is computed \emph{after} convergence, and a finalize policy $\pi_{\text{finalize}}$ synthesizes findings into a structured report (Appendix~\ref{appendix:finalize_prompt}).
Additional details on correctness properties, termination guarantees, and handling of inconclusive results are provided in Appendix~\ref{appendix:eog_properties}.

\subsection{Example scenario.}
\label{sec:example}

Consider a service chain: $S_1$ (frontend) $\to$ $S_2$ (gateway) $\to$ $S_3$ (processor) $\to$ $S_4$ (database). Crucially, \textbf{the edge $S_2 \to S_3$ is not in the topology graph}.$S_2$ calls $S_3$ dynamically based on request type, and this dependency was never registered. During a flash sale, traffic increases. Each hop amplifies the load slightly (retry storms, fan-out), and $S_4$, finally exhausts memory. The failure propagates back: $S_4 \to S_3 \to S_2$. An alert fires on $S_2$ (users see 500 errors).

\textbf{The diagnostic trap.} $S_4$'s Out Of Memory (OOM) error is \emph{strong local evidence} of being the origin. Without belief messages, the algorithm would terminate with $S_4$ as origin, leading operators to restart $S_4$ (without addressing the upstream traffic pattern). This example shows both \textbf{edge discovery} (finding the hidden $S_2 \to S_3$ dependency) and \textbf{belief revision} (correcting the $S_4$ misdiagnosis).
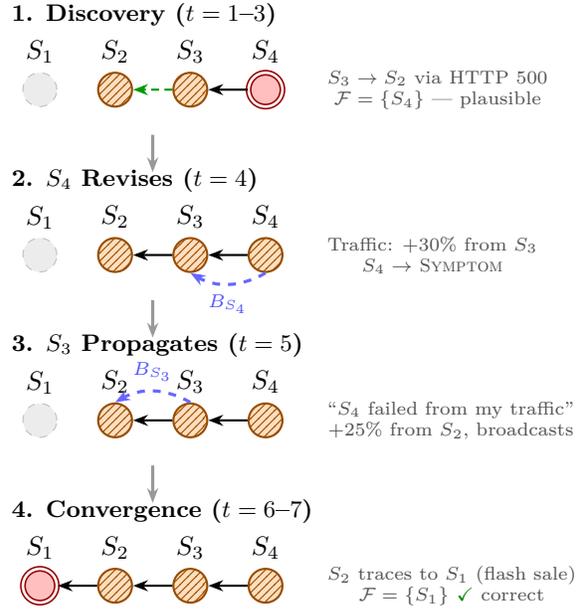
\begin{figure}[t]
    \centering
    \begin{tikzpicture}[
        node distance=0.8cm,
        gnode/.style={circle, draw, minimum size=0.45cm, font=\scriptsize\bfseries},
        origin/.style={gnode, fill=red!25, draw=red!60!black, double, double distance=1pt, line width=0.6pt},
        symptom/.style={gnode, fill=orange!25, draw=orange!60!black, line width=0.8pt,
            postaction={pattern=north east lines, pattern color=orange!60!black}},
        unknown/.style={gnode, fill=gray!15, draw=gray!50, dashed},
        arr/.style={-{Stealth[length=2mm]}, thick},
        discovered/.style={-{Stealth[length=2mm]}, thick, draw=green!60!black, densely dashed},
        msg/.style={-{Stealth[length=2mm]}, draw=blue!60, line width=1.2pt, dashed},
        evidence/.style={font=\scriptsize, text=black!70, align=center},
        phase/.style={font=\footnotesize\bfseries, align=left, anchor=west},
    ]

    \begin{scope}[local bounding box=p1]
        \node[phase] at (-2, 0.5) {1. Discovery ($t=1$--$3$)};
        \node[unknown, label=above:$S_1$] (s1a) at (-1.5, -0.5) {};
        \node[symptom, label=above:$S_2$] (s2a) at (-0.5, -0.5) {};
        \node[symptom, label=above:$S_3$] (s3a) at (0.5, -0.5) {};
        \node[origin, label=above:$S_4$] (s4a) at (1.5, -0.5) {};
        
        \draw[discovered] (s3a) -- (s2a);
        \draw[arr] (s4a) -- (s3a);
        \node[evidence, anchor=west] at (2.2, -0.5) {$S_3 \to S_2$ via HTTP 500\\$\mathcal{F}=\{S_4\}$ --- plausible};
    \end{scope}
    
    \draw[-{Stealth[length=2mm]}, very thick, draw=black!40] (0, -1.1) -- (0, -1.6);
    
    \begin{scope}[yshift=-2.2cm, local bounding box=p2]
        \node[phase] at (-2, 0.5) {2. $S_4$ Revises ($t=4$)};
        \node[unknown, label=above:$S_1$] (s1b) at (-1.5, -0.5) {};
        \node[symptom, label=above:$S_2$] (s2b) at (-0.5, -0.5) {};
        \node[symptom, label=above:$S_3$] (s3b) at (0.5, -0.5) {};
        \node[symptom, label=above:$S_4$] (s4b) at (1.5, -0.5) {};
        
        \draw[arr] (s3b) -- (s2b);
        \draw[arr] (s4b) -- (s3b);
        \draw[msg] (s4b.south) to[bend left=30] node[below, font=\scriptsize, text=blue!70] {$B_{S_4}$} (s3b.south);
        \node[evidence, anchor=west] at (2.2, -0.5) {Traffic: +30\% from $S_3$\\$S_4 \to$ \textsc{Symptom}};
    \end{scope}

    \draw[-{Stealth[length=2mm]}, very thick, draw=black!40] (0, -3.3) -- (0, -3.8);

    \begin{scope}[yshift=-4.4cm, local bounding box=p3]
        \node[phase] at (-2, 0.5) {3. $S_3$ Propagates ($t=5$)};
        \node[unknown, label=above:$S_1$] (s1c) at (-1.5, -0.5) {};
        \node[symptom, label=above:$S_2$] (s2c) at (-0.5, -0.5) {};
        \node[symptom, label=above:$S_3$] (s3c) at (0.5, -0.5) {};
        \node[symptom, label=above:$S_4$] (s4c) at (1.5, -0.5) {};
        
        \draw[arr] (s3c) -- (s2c);
        \draw[arr] (s4c) -- (s3c);
        \draw[msg] (s3c.north) to[bend right=30] node[above, font=\scriptsize, text=blue!70] {$B_{S_3}$} (s2c.north);
        \node[evidence, anchor=west] at (2.2, -0.5) {``$S_4$ failed from my traffic''\\+25\% from $S_2$, broadcasts};
    \end{scope}

    \draw[-{Stealth[length=2mm]}, very thick, draw=black!40] (0, -5.5) -- (0, -6.0);

    \begin{scope}[yshift=-6.6cm, local bounding box=p4]
        \node[phase] at (-2, 0.5) {4. Convergence ($t=6$--$7$)};
        \node[origin, label=above:$S_1$] (s1d) at (-1.5, -0.5) {};
        \node[symptom, label=above:$S_2$] (s2d) at (-0.5, -0.5) {};
        \node[symptom, label=above:$S_3$] (s3d) at (0.5, -0.5) {};
        \node[symptom, label=above:$S_4$] (s4d) at (1.5, -0.5) {};
        
        \draw[arr] (s2d) -- (s1d);
        \draw[arr] (s3d) -- (s2d);
        \draw[arr] (s4d) -- (s3d);
        \node[evidence, anchor=west] at (2.2, -0.5) {$S_2$ traces to $S_1$ (flash sale)\\$\mathcal{F}=\{S_1\}$ \textcolor{green!50!black}{\checkmark} correct};
    \end{scope}

    \end{tikzpicture}
    \vspace{-2mm}
    \caption{Edge discovery and belief revision sequence. (1) $S_2 \to S_3$ discovered. (2) $S_4$ revises belief due to high load. (3) $S_3$ propagates message. (4) $S_1$ identified as root cause.}
    \label{fig:belief-revision}\vspace{-9mm}
    \end{figure}

\begin{enumerate}[leftmargin=*,nosep]
    \item \textbf{$t=1$, Process $S_2$} \emph{(Alg.\ Steps 1--2, 4, 6)}: Alert node. CxC returns HTTP 500 errors showing failed calls to $S_3$. \textbf{Edge discovery}: $S_2 \to S_3$ was not in the topology graph, but traces reveal this dependency. The policy adds edge $S_3 \to S_2$ to $E_S$ and proposes $S_3$. $B_{S_2} = \textsc{Symptom}$ of $S_3$.
    
    \item \textbf{$t=2$, Process $S_3$}: Logs show timeouts calling $S_4$. Proposes $S_4$. $B_{S_3} = \textsc{Symptom}$ of $S_4$.
    
    \item \textbf{$t=3$, Process $S_4$}: Logs show \texttt{OutOfMemoryError}. \textbf{Strong local evidence of being the origin} points to memory exhaustion, that is  a classic root cause.
    
    At this point: $\mathcal{F} = \{S_4\}$. Single origin, plausible diagnosis. Without belief messages, \textbf{this would be the final step of the process} pointing to $S_4$'s memory issue.
    
    \item \textbf{$t=4$, $S_4$ re-enters ActiveSet} \emph{(Alg.\ Step 5: Propagation)}: CxC now includes traffic metrics (not just error logs). Policy sees: incoming request rate from $S_3$ is elevated (+30\% above baseline). Re-reasons: ``My OOM was caused by traffic overload, not internal failure.'' $B_{S_4} \to \textsc{Symptom}$ of $S_3$. \textbf{Broadcasts to $S_3$.}
    
    \item \textbf{$t=5$, $S_3$ receives the message} \emph{(Alg.\ Step 5: Propagation continues)}: Inbox contains $B_{S_4}$: ``I failed from \emph{your} traffic.'' $S_3$ checks its own incoming traffic: +25\% from $S_2$. Propagates message upstream. \textbf{Broadcasts to $S_2$.}
    
    \item \textbf{$t=6$, $S_2$ receives the message}: Inbox contains $B_{S_3}$: ``The cascade started from \emph{your} traffic.'' $S_2$ checks incoming: +20\% from $S_1$. Proposes $S_1$.
    
    \item \textbf{$t=7$, Process $S_1$}: Logs show flash sale triggered traffic spike. $B_{S_1} = \textsc{Origin}$.
    
    \item \textbf{Convergence} \emph{(Alg.\ Post-processing)}: $\mathcal{F} = \{S_1\}$---correct. The flash sale is the true origin.
\end{enumerate}

\textbf{Key insight.} This example demonstrates:
(1) \textbf{structure learning}: the hidden from topology graph $S_2 \to S_3$ edge was discovered from observability data~\cite{havrilla2025igda}; and (2) \textbf{belief revision}: $S_4$'s plausible OOM diagnosis was corrected via cascading messages. Without SBP, operators would restart $S_4$, providing temporary relief until memory fills again. The upstream traffic pattern is never addressed.

\section{Implementation \& Results}
\label{sec:results}
\begin{table}
\centering
\caption{MCP tools for SRE domain.}
\label{tab:mcp_tools}
\footnotesize
\begin{threeparttable}
\begin{tabular}{@{}lp{4.5cm}@{}}
\toprule
\textbf{Tool} & \textbf{Description} \\
\midrule
\texttt{alert\_summary} & Active alerts with severity \\
\texttt{get\_context\_contract} & Context Packet for entity \\
\texttt{topology\_analysis} & Dependency graph queries \\
\texttt{metric\_analysis} & Metrics with anomaly detection \\
\texttt{log\_analysis} & Service logs with filtering \\
\texttt{event\_analysis} & Events in time window \\
\texttt{get\_trace\_error\_tree} & Trace error propagation \\
\texttt{spec\_change\_analysis} & Config/deployment diffs \\
\bottomrule
\end{tabular}

\begin{tablenotes}[para]
\footnotesize
The ReAct agent has access to all tools, shell utilities, and can write Python code for data analysis. The EoG agent uses \texttt{alert\_summary}, \texttt{spec\_change\_analysis}, and \texttt{event\_analysis} to populate the initial ActiveSet, then invokes \texttt{get\_context\_contract} during entity visits where abductive policy is executed.
\end{tablenotes}
\end{threeparttable}
\vspace{-0.6cm}
\end{table}

\begin{table*}[]
\centering
\caption{EoG vs ReAct on ITBench (35 scenarios, 3 runs).}
\label{tab:rq1-results}
\resizebox{1.0\textwidth}{!}{%
\begin{threeparttable}
\begin{tabular}{ll|cccc|c|cc}
\hlineB{2}
& & \multicolumn{4}{c|}{\textbf{RC Entity}} & \textbf{RC Reas.} & & \\
\textbf{Agent} & \textbf{Model} & \textbf{pass@3 F1}$\uparrow$ & \textbf{pass@3 Rec}$\uparrow$ & \textbf{maj@3 F1}$\uparrow$ & \textbf{maj@3 Rec}$\uparrow$ & \textbf{maj@3 Rec}$\uparrow$ & \textbf{Input Tok.} & \textbf{Output Tok.} \\
\hlineB{2}
ReAct & GPT-5.1 & 22.9 & 22.9 & 8.6 & 11.4 & 10.0 & 779K & 9.3K \\
\cellcolor{blue!10} EoG & \cellcolor{blue!10} GPT-5.1 & \cellcolor{blue!10} \textbf{88.9} & \cellcolor{blue!10} \textbf{91.4} & \cellcolor{blue!10} \textbf{86.1} & \cellcolor{blue!10} \textbf{88.6} & \cellcolor{blue!10} \textbf{83.0} & \cellcolor{blue!10} 1837K & \cellcolor{blue!10} 1.7K \\
\hline
ReAct & GPT-OSS-120B & 28.6 & 28.6 & 5.7 & 5.7 & 8.6 & 230K & 3.8K \\
\cellcolor{blue!10} EoG & \cellcolor{blue!10} GPT-OSS-120B & \cellcolor{blue!10} \textbf{84.1} & \cellcolor{blue!10} \textbf{94.3} & \cellcolor{blue!10} \textbf{79.4} & \cellcolor{blue!10} \textbf{88.6} & \cellcolor{blue!10} \textbf{85.0} & \cellcolor{blue!10} 1751K & \cellcolor{blue!10} 2.1K \\
\hline
ReAct & Gemini 3 Flash & 88.6 & 94.3 & 74.3 & 74.3 & 72.9 & 5376K & 21.2K \\
\cellcolor{blue!20} \textbf{EoG} & \cellcolor{blue!20} \textbf{Gemini 3 Flash} & \cellcolor{blue!20} \textbf{94.5} & \cellcolor{blue!20} \textbf{97.1} & \cellcolor{blue!20} \textbf{92.1} & \cellcolor{blue!20} \textbf{94.3} & \cellcolor{blue!20} \textbf{90.2} & \cellcolor{blue!20} 3390K & \cellcolor{blue!20} 2.0K \\
\hline
ReAct & Kimi K2 & 48.6 & 48.6 & 28.6 & 28.6 & 22.9 & 1021K & 8.8K \\
\cellcolor{blue!10} EoG & \cellcolor{blue!10} Kimi K2 & \cellcolor{blue!10} \textbf{85.3} & \cellcolor{blue!10} \textbf{91.4} & \cellcolor{blue!10} \textbf{83.0} & \cellcolor{blue!10} \textbf{88.6} & \cellcolor{blue!10} \textbf{81.2} & \cellcolor{blue!10} 2641K & \cellcolor{blue!10} 1.9K \\
\hline
\cellcolor{blue!10} EoG & \cellcolor{blue!10} Minimax M2.1 & \cellcolor{blue!10} 72.9 & \cellcolor{blue!10} 88.6 & \cellcolor{blue!10} 64.2 & \cellcolor{blue!10} 77.1 & \cellcolor{blue!10} 68.0 & \cellcolor{blue!10} 1881K & \cellcolor{blue!10} 2.1K \\
\hline
\cellcolor{blue!10} EoG & \cellcolor{blue!10} Mistral Large & \cellcolor{blue!10} 76.6 & \cellcolor{blue!10} 91.4 & \cellcolor{blue!10} 67.2 & \cellcolor{blue!10} 80.0 & \cellcolor{blue!10} 76.6 & \cellcolor{blue!10} 1630K & \cellcolor{blue!10} 1.9K \\
\hlineB{2}
\end{tabular}
\begin{tablenotes}[para]
\footnotesize
Best overall shown in \colorbox{blue!20}{\textbf{bold}}; other EoG wins in \colorbox{blue!10}{light blue}. 
Token counts are averaged per trial per scenario and exclude reasoning tokens as providers do not consistently report them.
ReACT with GPT-OSS-20B, Mistral Large and Minimax M2.1 could not be run due to Codex failures.
\end{tablenotes}
\end{threeparttable}%
}
\end{table*}

\begin{table}[!ht]
\centering
\caption{ReAct prompted with EoG algorithm (Recall).}
\label{tab:rq2-prompting}
\begin{threeparttable}
\begin{tabular}{l|cc}
\hlineB{2}
\textbf{Model} & \textbf{pass@3}$\uparrow$ & \textbf{maj@3}$\uparrow$ \\
\hlineB{2}
GPT-5.1 & 40.0 & 2.9 \\
Gemini 3 Flash & 71.4 & 0.0 \\
Kimi K2 & 42.9 & 0.0 \\
\hlineB{2}
\end{tabular}
\begin{tablenotes}[para]
\footnotesize
ReAct agent prompted with EoG algorithm instructions. Compare with EoG results in Table~\ref{tab:rq1-results} (e.g., GPT-5.1: 91.4/88.6 vs.\ 40.0/2.9 here).
\end{tablenotes}
\end{threeparttable}
\end{table}

\begin{table}[!ht]
\centering
\caption{SBP ablation (maj@3 Recall, RC Entity).}
\label{tab:rq3-ablation}
\begin{threeparttable}
\begin{tabular}{l|c|c}
\hlineB{2}
\textbf{Model} & \textbf{EoG-SBP} & \textbf{$\Delta$SBP}$\uparrow$ \\
\hlineB{2}
GPT-5.1 & 82.9 & \cellcolor{blue!10} +6.9\% \\
GPT-OSS-120B & 74.3 & \cellcolor{blue!10} +19.2\% \\
GPT-OSS-20B & 40.0 & \cellcolor{blue!20} \textbf{+64.3\%} \\
Gemini 3 Flash & 85.7 & \cellcolor{blue!10} +10.0\% \\
Kimi K2 & 77.1 & \cellcolor{blue!10} +14.9\% \\
Minimax M2.1 & 60.0 & \cellcolor{blue!10} +28.5\% \\
Mistral Large & 71.4 & \cellcolor{blue!10} +12.0\% \\
\hlineB{2}
  \end{tabular}
\begin{tablenotes}[para]
\footnotesize
EoG-SBP = Deterministic Controller only. $\Delta$SBP = relative improvement from adding SBP. Largest gain in \colorbox{blue!20}{\textbf{bold}}.
  \end{tablenotes}
\end{threeparttable}
\vspace{-0.6cm}
\end{table}

\paragraph{Implementation.} We implement the EoG agent and a baseline ReAct agent on the Codex agent framework,\footnote{Available at \url{https://github.com/openai/codex}} and evaluate on abductive reasoning tasks in ITBench (\S\ref{sec:motivation}). The implementation comprises \textasciitilde{5,000} lines of Rust code. The ReAct baseline incorporates best practices from prior work on LLM-based diagnostics~\cite{roy2024exploring}, achieving higher performance than published baselines on ITBench; implementation details are provided in Appendices~\ref{appendix:react_agent} and~\ref{appendix:eog_implementation}.

We implement domain-specific MCP servers exposing the operational graph and evidence data (e.g., logs, traces, and metrics in ITBench), including distributed traces~\cite{kaldor2017canopy}, via the tools in Table~\ref{tab:mcp_tools}.

For each entity $v$, the Context Packet satisfies the Local Markov Property: it contains entity metadata, 1-hop neighbors, time-windowed observability data, and the current Inbox. Large contexts ($>$80K tokens) are chunked with overlap and classified via map-reduce. Domain expertise is encoded in structured prompts organized by phase (bootstrap, explore, finalize), with the core algorithm logic remaining domain-agnostic.

\paragraph{Experiments.}
35 ITBench scenarios spanning misconfigurations, resource exhaustion, and cascading failures, are used for evaluation, with 3 runs per scenario to measure both accuracy and consistency. All experiments use high reasoning effort settings. Notably, ReAct has access to \emph{more} tools (all MCP tools plus shell utilities and Python code execution) while EoG uses only the Context Contract per entity, and controls the investigation structure algorithmically.

We compute two metrics: \textbf{RC Entity}: a correctness of identified root cause entity (e.g., \texttt{otel-demo/Pod/cartservice-xyz}), and \textbf{RC Reasoning}: a correctness of explanation of the causal mechanism (e.g., ``Pod crashed due to incorrect OS architecture in container image''). More details in ~\ref{appendix:evaluation}.

\textbf{EoG vs ReAct.} Table~\ref{tab:rq1-results} compares EoG against ReAct baselines. The results directly address the \emph{reliability gap} identified in \S\ref{sec:motivation-variance}: while ReAct exhibits a large variance between pass@3 and majority@3 (indicating stochastic success), EoG substantially narrows this gap. For example, GPT-5.1 under ReAct achieves only 22.9\% pass@3 and 8.6\% majority@3 F1 (RC Entity); under EoG, these rise to 88.9\% and 86.1\% respectively, near-closing the reliability gap.

Additionally, ablation experiments of removing distributed traces as a data source showed no statistically significant change in results; this is consistent with the redundancy typical of observability data, where the same diagnostic evidence can often be inferred from multiple signals (logs, metrics, events).

\textbf{Prompting Ablation.} We quantify the efficacy of a ReAct agent prompted to follow the EoG algorithm (\S\ref{sec:algorithm}) without the external Deterministic Controller. Table~\ref{tab:rq2-prompting} shows that majority@3 collapses to near zero despite non-trivial pass@3, indicating that the cognitive load of simultaneously managing exploration state, message passing, and causal reasoning overwhelms instruction-following capacity.

\textbf{SBP Ablation.} Table~\ref{tab:rq3-ablation} isolates the contribution of SBP: structured traversal alone captures most gains over ReAct, while SBP provides 7--64\% additional improvement by enabling post-hoc belief correction (\S\ref{sec:motivation-ordering}), with the largest gains on weaker models where early misclassifications are more common.

\section{Related Work}
\label{sec:related}
\begin{table}[t]
\centering
\caption{Comparison of LLM/agent baselines.}
\label{tab:baseline-comparison}
\setlength{\tabcolsep}{2pt}
\begin{tabular}{@{}l|ccccc|c@{}}
\toprule
& \rotatebox{90}{\small ReAct\,\cite{yao2023react}~~} 
& \rotatebox{90}{\small Graph-RAG\,\cite{edge2024graphrag}~~} 
& \rotatebox{90}{\small RCA\,\cite{roy2024exploring}~~} 
& \rotatebox{90}{\small STRATUS\,\cite{chen2025stratus}~~} 
& \rotatebox{90}{\small GALA\,\cite{tian2025gala}~~} 
& \rotatebox{90}{\small\textbf{EoG (Ours)}~~} \\
\midrule
{\small Graph as mutable state} & {\scriptsize\ding{55}} & {\scriptsize$\circ$} & {\scriptsize$\circ$} & {\scriptsize$\circ$} & {\scriptsize$\circ$} & {\scriptsize\ding{51}} \\
{\small Hypotheses explicit}    & {\scriptsize\ding{55}} & {\scriptsize\ding{55}} & {\scriptsize$\circ$} & {\scriptsize$\circ$} & {\scriptsize$\circ$} & {\scriptsize\ding{51}} \\
{\small Belief revision}        & {\scriptsize\ding{55}} & {\scriptsize\ding{55}} & {\scriptsize\ding{55}} & {\scriptsize\ding{55}} & {\scriptsize\ding{55}} & {\scriptsize\ding{51}} \\
{\small Propagation}            & {\scriptsize glob} & {\scriptsize g.sum} & {\scriptsize glob} & {\scriptsize glob} & {\scriptsize hyb} & {\scriptsize msg} \\
{\small Controller invariants}  & {\scriptsize\ding{55}} & {\scriptsize\ding{55}} & {\scriptsize$\circ$} & {\scriptsize$\circ$} & {\scriptsize$\circ$} & {\scriptsize\ding{51}} \\
{\small Data dynamicity}        & {\scriptsize epis} & {\scriptsize epis} & {\scriptsize str} & {\scriptsize str} & {\scriptsize str} & {\scriptsize str} \\
\bottomrule
\end{tabular}
\\[2pt]
{\scriptsize \ding{51}=yes, \ding{55}=no, $\circ$=partial. epis=episodic, str=streaming.}\\
{\scriptsize Propagation: how updates spread---glob=entire context passed each step,}
{\scriptsize g.sum=global with summarization, hyb=mix of global/local, msg=local message passing.}
\end{table}

Our work relates to LLM-based investigative agents performing multi-step reasoning over structured environments~\cite{yao2023react, bei2025graphs, roy2024exploring, xu2025openrca, chen2024automatic, tian2025gala, sun2024thinkongraph, luo2024rog}. Recent work introduces belief-centric abstractions~\cite{lidayan2025abbel, kim2024qube, he2025enhancing}, but treats beliefs as prompt-level artifacts rather than mutable state. These systems assume monotonic updates and lack explicit mechanisms for belief revision in presence of contradictory evidence.
Causal discovery and effect estimation~\cite{wan2024large, vowels2022d, yao2021survey} while related, target different outputs under static, fully-observed data and do not model revisable explanatory hypotheses under streaming regimes.

Graph-RAG systems~\cite{edge2024graphrag, he2024gretriever, liu2025polyg, markovic2025optimizing, kg-injection-2025, besta2025affordable, sun2024thinkongraph, luo2024rog, jiang2023structgpt} enable multi-hop reasoning over knowledge graphs. Theorem-of-Thought~\cite{abdaljalil2025theorem} uses Bayesian belief propagation over parallel agent reasoning graphs to aggregate multiple reasoning attempts; in contrast, EoG uses message passing to correct sequential mistakes as new evidence emerges. 

Diagnostic agents for cloud operations~\cite{roy2024exploring, xu2025openrca, zhang2025tamo, deshpande2025trailtracereasoningagentic, chen2025stratus, tian2025gala, ma2025diagnosingfailurerootcauses} generate explanatory hypotheses from partial evidence. Graph-based systems~\cite{wang2021groot, tonon2025radice, zhu2024root, tang2025root, liu2025graphlocator, xiang2025simplifying} use dependency graphs to localize faults. All these rely on \emph{append-only} inference without explicit revision or localized propagation, making them brittle under evolving evidence in unbounded graphs.

Table~\ref{tab:baseline-comparison} summarizes the comparison of EOG with representative ReAct, Graph-RAG, and diagnostic agents, and ablate revision and propagation, demonstrating that gains arise from state semantics rather than domain-specific engineering.

\section{Conclusion}
Diagnostic investigations over massive, heterogeneous data with hidden dependency structure require consistent correctness; stochastic success is insufficient when verification carries risk. We introduced EoG, a disaggregated architecture that separates local abductive reasoning via LLMs from deterministic graph traversal and belief propagation, achieving $7\times$ higher Majority@k F1 on ITBench scenarios. This separation of concerns generalizes to other domains where explanations must be derived from environments with inherent graphical structure.

\newpage
\section{Impact Statement}
This paper presents work whose goal is to advance the field of machine learning. There are many potential societal consequences of our work, none of which we feel must be specifically highlighted here.

\newpage

\bibliography{main}
\bibliographystyle{icml2026}

\appendix

\section{ReAct Agent Implementation Details}
\label{appendix:react_agent}

We provide implementation details for the ReAct baseline agent~\cite{yao2023react} used in our evaluation. The baseline leverages OpenAI Codex CLI (v0.76)~\cite{openai_codex}, an open-source coding agent that runs locally and supports extensible tool integration via the Model Context Protocol (MCP).

\subsection{OpenAI Codex CLI Architecture}
\label{appendix:codex_architecture}

\begin{figure}[h]
\centering
\begin{tikzpicture}[
    node distance=0.5cm and 0.8cm,
    box/.style={draw, rounded corners=3pt, minimum height=0.55cm, font=\scriptsize, align=center},
    llm/.style={box, fill=orange!15, draw=orange!60!black, minimum width=2cm},
    tool/.style={box, fill=green!10, draw=green!60!black, minimum width=1.4cm},
    mcp/.style={box, fill=blue!10, draw=blue!60!black, minimum width=1.4cm},
    sandbox/.style={box, fill=gray!10, draw=gray!60!black, minimum width=1.6cm},
    arr/.style={-{Stealth[length=1.5mm]}, thick},
    phase/.style={font=\scriptsize\bfseries, text=black!80},
]

\node[llm, minimum width=2.5cm, minimum height=0.7cm] (core) at (0, 2) {Codex CLI Core};
\node[phase] at (0, 2.7) {OpenAI Codex CLI (v0.76)};

\node[tool] (files) at (-2.8, 0.5) {File Ops};
\node[tool] (shell) at (-1, 0.5) {Shell};
\node[tool] (python) at (0.8, 0.5) {Python};
\node[tool] (summ) at (2.6, 0.5) {Summarize};

\node[mcp, minimum width=5cm] (mcp) at (0, -0.7) {Model Context Protocol (MCP)};

\node[sandbox, fill=red!8, draw=red!50!black] (sre) at (-1.6, -2) {SRE Tools};
\node[sandbox, fill=purple!8, draw=purple!50!black] (custom) at (1.6, -2) {Custom MCP};

\draw[arr] (core) -- (files);
\draw[arr] (core) -- (shell);
\draw[arr] (core) -- (python);
\draw[arr] (core) -- (summ);
\draw[arr] (files) -- (mcp);
\draw[arr] (shell) -- (mcp);
\draw[arr] (python) -- (mcp);
\draw[arr] (summ) -- (mcp);
\draw[arr] (mcp) -- (sre);
\draw[arr] (mcp) -- (custom);

\node[font=\tiny, text=black!60, anchor=west] at (3.3, 2) {LLM Loop};
\node[font=\tiny, text=black!60, anchor=west] at (3.3, 0.5) {Built-in};
\node[font=\tiny, text=black!60, anchor=west] at (3.3, -0.7) {Interface};
\node[font=\tiny, text=black!60, anchor=west] at (3.3, -2) {Extensions};

\end{tikzpicture}
\caption{OpenAI Codex CLI architecture. The core agent loop interfaces with built-in tools through the Model Context Protocol. Domain-specific MCP servers extend capabilities.}
\label{fig:codex-architecture}
\end{figure}
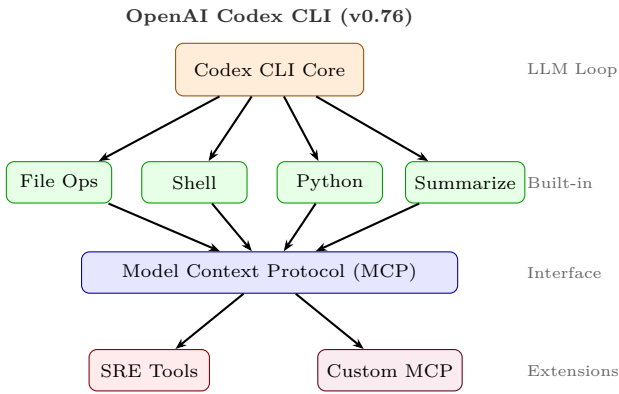

Codex CLI implements a \textbf{disaggregated architecture} where the LLM reasoning loop is separated from tool execution~\cite{openai_codex}. The core is written in Rust (\texttt{codex-rs}) and provides:

\begin{itemize}[nosep]
    \item \textbf{Sandboxed Execution}: Shell and code execution occur in isolated subprocesses with configurable permission levels (read-only, write workspace, full access). The sandbox restricts filesystem writes, network access, and system modifications by default.
    \item \textbf{MCP Tool Servers}: Tools run as separate processes communicating via the Model Context Protocol. This separation enables: (1) crash isolation---a failing tool does not crash the agent; (2) resource limits---each tool server can be independently constrained; (3) extensibility---domain tools are added without modifying the core.
    \item \textbf{Automatic Context Summarization}: When tool outputs exceed context limits, Codex summarizes content to preserve important information while staying within token bounds.
    \item \textbf{State Persistence}: Python execution maintains state across invocations; file operations are transactional within the workspace.
\end{itemize}

\subsection{ReAct Agent Prompt}
\label{appendix:react_prompt}

The ReAct baseline operates as a Site Reliability Engineering (SRE) support agent designed for offline incident analysis. The agent follows a structured investigation workflow with explicit phases for context discovery, symptom analysis, hypothesis generation, evidence collection, and causal chain construction. The full system prompt is provided below, with environment variables (\texttt{\$SNAPSHOT\_DIRS}, \texttt{\$WORKSPACE\_DIR}) populated at runtime.

\begin{promptbox}{ReAct with Code System Prompt}
\scriptsize
\textbf{Task:} You are an expert SRE (Site Reliability Engineer) and Kubernetes SRE Support Agent investigating a production incident from OFFLINE snapshot data.

You are a highly capable tool-using agent able to: Diagnose Kubernetes failures; Correlate alerts, events, traces, and metrics; Identify contributing factors (not just root cause); Perform data analysis using Python when useful.

\vspace{0.3em}
\textbf{INCIDENT SNAPSHOT DATA LOCATION}\\
Your incident data is located in \texttt{\$SNAPSHOT\_DIRS} (READ-ONLY). Do NOT search the filesystem for additional data. Start by listing the contents of these directories. Working directory: \texttt{\$WORKSPACE\_DIR}. Application topology available at \texttt{\$WORKSPACE\_DIR/app.json}.

\vspace{0.3em}
\textbf{FINAL OUTPUT FORMAT (MANDATORY)}\\
Generate a JSON diagnosis identifying all Kubernetes entities: entities that CAUSED the incident (\texttt{contributing\_factor=true}), entities IMPACTED but did not cause it (\texttt{contributing\_factor=false}), and the propagation chain showing how the incident spread. Requirements: explain all firing alerts, provide reasoning/evidence for every entity, construct fault propagation chain, use Python for data analysis when necessary. Write diagnosis to \texttt{\$WORKSPACE\_DIR/agent\_output.json}. Validate JSON using \texttt{jq} after writing.

\vspace{0.3em}
\textbf{ENTITY NAMING CONVENTION}\\
All entities MUST use format: \texttt{namespace/Kind/name}. Examples: \texttt{otel-demo/Deployment/ad}, \texttt{otel-demo/Service/frontend}. Do NOT include UIDs.

\vspace{0.3em}
\textbf{PROPAGATION CHAIN (MANDATORY)}\\
Construct chain: Root Cause $\to$ Intermediate Services $\to$ Impacted Services. Each link includes: \texttt{source} (causing entity), \texttt{target} (affected entity), \texttt{condition} (what state caused propagation), \texttt{effect} (observable effect on target).

\vspace{0.3em}
\textbf{RULES FOR CONTRIBUTING FACTORS}\\
\texttt{contributing\_factor=true} (IRREDUCIBLE): Mark ONLY if entity is an independent cause not fully explained by another contributing factor. Use ``irreducibility test'': if failure is explained by ``upstream X failed'' with propagation edge $X \to$ entity, then entity is NOT irreducible $\to$ set \texttt{false}.\\
\texttt{contributing\_factor=false} (DERIVED): Mark entities that are downstream effects/symptoms caused by another contributing factor.\\
IMPORTANT: Do NOT mark both cause and derived symptom as \texttt{true}. Multiple \texttt{contributing\_factors} allowed ONLY if truly independent.

\vspace{0.3em}
\textbf{DEBUGGING PRINCIPLES}\\
(1) Differential Observability: Compare replicas and time windows.
(2) Occam's Razor: Choose simplest explanation consistent with evidence.
(3) Duration Matching: Theory must explain entire incident duration.
(4) Follow Breadcrumbs: Let alerts/errors guide investigation.
(5) No Jumping to Conclusions: Validate every hypothesis with evidence.
(6) Chaos Files $\neq$ Active Chaos: Verify experiment was running AND time-aligned.
(7) Semantic Name Normalization: Try variations, strip suffixes, search partial matches.

\vspace{0.3em}
\textbf{INVESTIGATION WORKFLOW (DO NOT SKIP)}\\
\textbf{Phase 1 -- Context Discovery:} List available files; read topology.md to understand dependencies.\\
\textbf{Phase 2 -- Symptom Analysis:} Read all alert files; compute start/end time, duration, frequency; build Alerts Table.\\
\textbf{Phase 3 -- Hypothesis Generation:} Create initial hypotheses; create validation plan for each.\\
\textbf{Phase 4 -- Evidence Collection Loop:} Use tools and Python to gather log/event/metrics/trace evidence; validate/refute hypotheses; explain firing alerts.\\
\textbf{Phase 5 -- Causal Chain Construction:} Build chain: [Config Error] $\to$ [CrashLoop] $\to$ [Service Down] $\to$ [Frontend 5xx].\\
\textbf{Phase 6 -- Conclusion:} Ensure all alerts explained, all entities included, contributions correctly labeled; output final JSON.

\vspace{0.3em}
\textbf{PYTHON ANALYSIS (ENCOURAGED)}\\
Write Python snippets to: parse alerts/events/metrics/traces, join datasets by timestamp or UID, extract failing spans, compute metrics deltas, identify patterns (CrashLoopBackOff, OOMKilled, 5xx spikes). Persist snippets in Python files.

\vspace{0.3em}
\textbf{PROHIBITED ACTIONS}\\
NEVER read ground\_truth.yaml; NEVER output anything other than final JSON; NEVER hallucinate K8s objects; NEVER leave alerts unexplained.
\end{promptbox}

\subsection{Expected Output Schema}
\label{appendix:output_schema}

Both the ReAct baseline and EoG agent must produce a structured JSON diagnosis with three key components:

\begin{lstlisting}[basicstyle=\scriptsize\ttfamily,frame=single]
{
  "entities": [{
    "name": "namespace/Kind/name",
    "contributing_factor": true|false,
    "reasoning": "Explanation for involvement",
    "evidence": "Supporting facts summary"
  }],
  "propagations": [{
    "source": "namespace/Kind/source-name",
    "target": "namespace/Kind/target-name",
    "condition": "What caused propagation",
    "effect": "What effect was observed"
  }],
  "alerts_explained": [{
    "alert": "<alert name>",
    "explanation": "Why the alert fired",
    "explained": true|false
  }]
}
\end{lstlisting}

\subsection{SRE Domain Tools (MCP Server)}
\label{appendix:sre_tools}

We implement a custom MCP server providing specialized SRE investigation tools. Table~\ref{tab:sre_tools_full} provides detailed function signatures for the primary tools.

\begin{table*}[t]
\centering
\caption{SRE Utility Tools Available via MCP}
\label{tab:sre_tools_full}
\small
\begin{tabular}{p{4cm}p{12cm}}
\toprule
\textbf{Tool} & \textbf{Description} \\
\midrule
\texttt{build\_topology} & Build operational topology graph from application architecture and Kubernetes objects \\
\texttt{topology\_analysis} & Analyze topology for dependencies, service context, infrastructure hierarchy \\
\texttt{metric\_analysis} & Analyze metrics with filtering, grouping, derived metrics (SQL-like interface) \\
\texttt{get\_metric\_anomalies} & Detect anomalies in time-series metrics \\
\texttt{event\_analysis} & Analyze K8s events with filtering, grouping, aggregation \\
\texttt{get\_trace\_error\_tree} & Analyze distributed traces with error trees and latency percentiles \\
\texttt{alert\_summary} & High-level alert overview: type, entity, duration, frequency \\
\texttt{alert\_analysis} & Detailed alert analysis with filtering and grouping \\
\texttt{k8s\_spec\_change\_analysis} & Track K8s object spec changes over time \\
\texttt{get\_k8\_spec} & Retrieve K8s spec for a specific resource \\
\texttt{get\_context\_contract} & Aggregated context for an entity (events, alerts, traces, metrics, spec changes) \\
\bottomrule
\end{tabular}
\end{table*}

\paragraph{Operational Graph Construction.}
The \texttt{build\_topology} tool constructs the operational graph $G$ by combining static information (application architecture, service dependencies declared in configuration) with dynamic runtime state (e.g., pod-to-node bindings, replica counts, endpoint mappings from Kubernetes). Such operational graphs are readily available through commercial observability platforms---including IBM Instana~\cite{ibminstana}, Dynatrace~\cite{dynatrace}, and Datadog~\cite{datadog}---which automatically discover and maintain service topologies through agent instrumentation and API introspection.

\paragraph{Key Tool Signatures.}

\textbf{\texttt{metric\_analysis}}: SQL-like interface for metric analysis supporting batch queries, derived metrics, and flexible grouping.
\begin{lstlisting}[language=Python,basicstyle=\scriptsize\ttfamily,frame=single]
def metric_analysis(
    base_dir: str,              # Metrics directory path
    k8_object_name: str = None, # Optional: specific object
    object_pattern: str = "*",  # Glob pattern (e.g., 'pod/*')
    metric_names: List[str] = [],  # Metrics to analyze
    eval: str = None,           # Derived metric expression
    group_by: str = None,       # Grouping column
    agg: str = "mean",          # Aggregation function
    start_time: str = None,     # Time range start
    end_time: str = None        # Time range end
) -> dict
\end{lstlisting}

\textbf{\texttt{get\_trace\_error\_tree}}: Analyzes distributed traces to find critical paths with regressions.
\begin{lstlisting}[language=Python,basicstyle=\scriptsize\ttfamily,frame=single]
def get_trace_error_tree(
    trace_file: str,            # Path to traces TSV
    service_name: str = None,   # Filter by service
    pivot_time: str = None,     # Timestamp for before/after
    delta_time: str = "5m",     # Comparison window size
    error_threshold_pct: float = 10,   # Error change threshold
    latency_threshold_pct: float = 10  # Latency change threshold
) -> dict
\end{lstlisting}

\textbf{\texttt{get\_context\_contract}}: The primary tool for EoG, returning full operational context for an entity.
\begin{lstlisting}[language=Python,basicstyle=\scriptsize\ttfamily,frame=single]
def get_context_contract(
    k8_object: str,             # K8s object (Kind/name format)
    snapshot_dir: str,          # Snapshot directory path
    topology_file: str = None,  # Optional topology JSON path
    start_time: str = None,     # Time range start
    end_time: str = None,       # Time range end
    page: int = 1,              # Page number
    deps_per_page: int = 3      # Dependencies per page
) -> dict
\end{lstlisting}

\section{EoG Implementation Details}
\label{appendix:eog_implementation}

This section provides complete architectural and prompt specifications for the EoG (Explanations over Graphs) agent, sufficient for reproduction.

\subsection{Architecture Overview}
\label{appendix:eog_arch_overview}

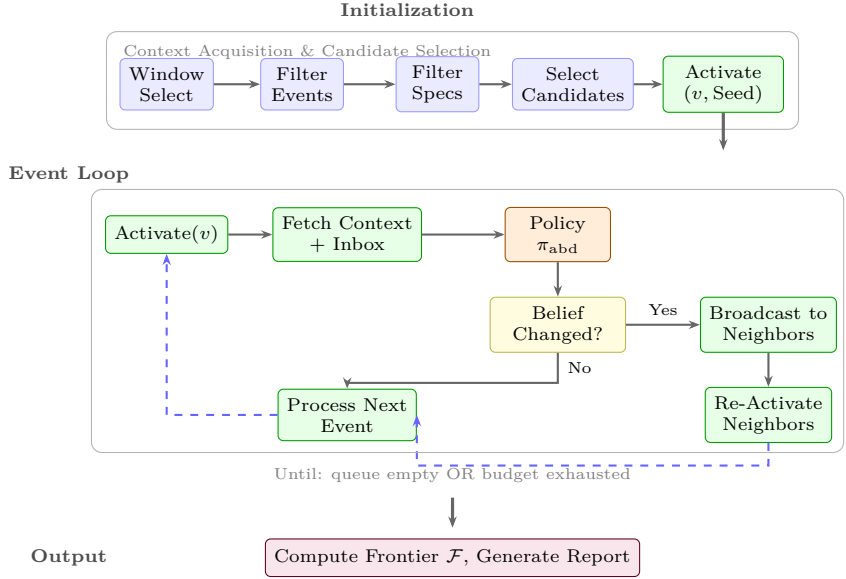
\begin{figure*}[t]
\centering
\begin{tikzpicture}[
    node distance=0.4cm and 0.6cm,
    box/.style={draw, rounded corners=2pt, minimum height=0.5cm, font=\scriptsize, align=center},
    phase/.style={font=\scriptsize\bfseries, text=black!70},
    stepbox/.style={box, fill=blue!8, draw=blue!40, minimum width=1.1cm},
    llmbox/.style={box, fill=orange!15, draw=orange!60!black},
    ctrlbox/.style={box, fill=green!10, draw=green!60!black},
    decbox/.style={box, fill=yellow!15, draw=yellow!70!black, minimum width=1.8cm},
    outbox/.style={box, fill=purple!10, draw=purple!50!black},
    arr/.style={-{Stealth[length=1.5mm]}, thick, draw=black!60},
    looparr/.style={-{Stealth[length=1.5mm]}, thick, draw=blue!60, dashed},
    container/.style={draw=black!30, rounded corners=4pt, inner sep=6pt},
]

\node[phase] (init-label) at (0, 4.8) {Initialization};

\node[stepbox] (window) at (-3.2, 3.8) {Window\\Select};
\node[stepbox] (filter-evt) at (-1.4, 3.8) {Filter\\Events};
\node[stepbox] (filter-spec) at (0.4, 3.8) {Filter\\Specs};
\node[stepbox] (bootstrap) at (2.2, 3.8) {Select\\Candidates};
\node[ctrlbox, minimum width=1.6cm] (seed) at (4.2, 3.8) {Activate\\$(v, \text{Seed})$};

\draw[arr] (window) -- (filter-evt);
\draw[arr] (filter-evt) -- (filter-spec);
\draw[arr] (filter-spec) -- (bootstrap);
\draw[arr] (bootstrap) -- (seed);

\draw[black!30, rounded corners=4pt] (-4.0, 3.2) rectangle (5.2, 4.5);
\node[font=\tiny, text=black!50, anchor=north west] at (-3.9, 4.45) {Context Acquisition \& Candidate Selection};

\draw[arr, very thick] (seed.south) -- ++(0, -0.5);

\node[phase] (loop-label) at (-4.5, 2.6) {Event Loop};

\node[ctrlbox, minimum width=1.4cm] (activate) at (-3.2, 1.8) {Activate$(v)$};
\node[ctrlbox, minimum width=1.8cm] (fetch) at (-0.8, 1.8) {Fetch Context\\+ Inbox};
\node[llmbox, minimum width=1.4cm] (policy) at (2.0, 1.8) {Policy\\$\pi_{\text{abd}}$};

\draw[arr] (activate) -- (fetch);
\draw[arr] (fetch) -- (policy);

\node[decbox] (decision) at (2.0, 0.6) {Belief\\Changed?};
\draw[arr] (policy) -- (decision);

\node[ctrlbox, minimum width=1.6cm] (broadcast) at (4.8, 0.6) {Broadcast to\\Neighbors};
\node[ctrlbox, minimum width=1.6cm] (reactivate) at (4.8, -0.6) {Re-Activate\\Neighbors};

\draw[arr] (decision.east) -- node[above, font=\tiny] {Yes} (broadcast.west);
\draw[arr] (broadcast) -- (reactivate);

\node[ctrlbox, minimum width=1.8cm] (next) at (-0.8, -0.6) {Process Next\\Event};

\draw[arr] (decision.south) -- node[right, font=\tiny] {No} ++(0, -0.4) -| (next.north);
\draw[looparr] (reactivate.south) -- ++(0, -0.3) -| (next.east);
\draw[looparr] (next.west) -| (activate.south);

\node[font=\tiny, text=black!50, align=center] at (0.6, -1.4) {Until: queue empty OR budget exhausted};

\draw[black!30, rounded corners=4pt] (-4.2, -1.1) rectangle (5.8, 2.4);

\draw[arr, very thick] (0.6, -1.7) -- ++(0, -0.4);

\node[phase] (out-label) at (-4.5, -2.5) {Output};
\node[outbox, minimum width=4cm] (output) at (0.6, -2.5) {Compute Frontier $\mathcal{F}$, Generate Report};

\end{tikzpicture}
\caption{EoG implementation architecture. \textbf{Initialization}: Multi-phase context acquisition filters events and spec changes, then bootstraps initial candidates. \textbf{Event Loop}: Each \texttt{Activate} event triggers context fetch, policy inference, and conditional belief broadcast. The loop continues until the queue empties or budget exhausts. \textbf{Output}: The frontier $\mathcal{F}$ is computed and a structured report is generated.}
\label{fig:eog-impl-architecture}
\end{figure*}

Figure~\ref{fig:eog-impl-architecture} illustrates the complete EoG execution flow, from initialization through the event-driven inference loop to final output generation.

\subsection{Event-Driven Architecture}
\label{appendix:event_architecture}

The implementation realizes the ActiveSet abstraction from \S\ref{sec:algorithm} as an event-driven message queue. This architectural choice enables concurrent node evaluation without changing the algorithm's semantics---each event type corresponds directly to operations in the abstract algorithm.

\begin{table*}[t]
\centering
\caption{Event Types and Semantics}
\label{tab:event_types}
\small
\begin{tabular}{lp{6cm}}
\toprule
\textbf{Event} & \textbf{Semantics} \\
\midrule
\texttt{Activate(v, reason)} & Evaluate node $v$; equivalent to \texttt{ActiveSet.pop(v)} \\
\texttt{Message(u, v, $B_u$, t)} & Deliver belief $B_u$ to $v$'s inbox \\
\texttt{BeliefChanged(v)} & Broadcast $B_v$ to neighbors; triggers re-activation \\
\bottomrule
\end{tabular}
\end{table*}

The event queue processes \texttt{Activate} events by: (1) checking visit eligibility, (2) collecting inbox messages, (3) fetching local context, (4) invoking the abductive policy, (5) applying damping if oscillating, (6) emitting \texttt{BeliefChanged} if belief changed, and (7) emitting \texttt{Activate} for discovered or proposed nodes. This is semantically equivalent to the ActiveSet-based loop but enables the actor model for concurrent processing.

\subsection{Termination Safeguards}
\label{appendix:termination}

Beyond the damping heuristic ($k_{\text{thresh}}$) described in \S\ref{sec:algorithm}, the implementation adds visit limits for robustness:

\begin{itemize}[nosep]
    \item \textbf{Max Visits}: Each node may be evaluated at most $k_{\text{max}}$ times (default: 5), regardless of belief stability.
    \item \textbf{Cooldown}: Re-evaluation requires at least $k_{\text{cool}}$ other evaluations since last visit (default: 2), preventing rapid re-activation loops.
\end{itemize}

Together with damping, these bound total inference to $O(\min(k_{\text{thresh}}, k_{\text{max}}) \cdot |V_S|)$.

\subsection{Checkpointing}
\label{appendix:checkpointing}

The implementation supports fault-tolerant checkpointing at quiescent points between event processing. Checkpoints capture node beliefs, inbox contents, pending events, and the explanatory graph $(V_S, E_S)$, enabling recovery without restarting the investigation.

\subsection{Initialization Prompts}
\label{appendix:init_prompts}

The investigation begins with a multi-phase initialization that sets the algorithm variables from \S\ref{sec:algorithm}:
\texttt{select\_investigation\_window} determines time window $W$;
\texttt{filter\_events} and \texttt{filter\_spec\_changes} acquire bounded context;
\texttt{bootstrap} selects initial candidates $V_\mathcal{O}$ for the ActiveSet.

\subsubsection{Investigation Window Selection}

The first step determines the temporal scope by identifying ``golden signal'' alerts.

\begin{promptbox}{select\_investigation\_window}
\scriptsize
\texttt{You are an SRE agent. Your goal is to choose an investigation time window [start\_time, end\_time] that best captures the onset and peak of the incident, using golden signal alerts as anchors.}

\vspace{0.3em}
\texttt{Golden signals include alerts indicating: Errors (request error rate, 5xx), Latency (p95/p99), Traffic (throughput drop, no requests).}

\vspace{0.3em}
\texttt{Ignore noisy platform alerts (watchdog, inhibitors) unless they directly explain golden-signal symptoms.}

\vspace{0.3em}
\texttt{\textbf{Input:} Alert Summary (JSON with names, entities, timestamps, severity), Application Architecture (service dependencies).}

\vspace{0.3em}
\texttt{\textbf{Output:} JSON with start\_time, end\_time, anchor\_alerts, golden\_signal\_alerts, reasoning.}

\vspace{0.3em}
\texttt{\textbf{Rules:} Prefer application services over infrastructure alerts. start\_time $\leq$ end\_time. Use timestamps from alert summary.}
\end{promptbox}

\subsubsection{Event Filtering}

Kubernetes clusters generate thousands of events per hour. We filter events in pages, selecting those relevant to the investigation.

\begin{promptbox}{filter\_events}
\scriptsize
\texttt{You are filtering Kubernetes events to identify those relevant to an ongoing incident investigation.}

\vspace{0.3em}
\texttt{\textbf{Relevance Criteria --- INCLUDE:}}
\begin{itemize}[nosep,leftmargin=1em]
    \item \texttt{Warning/Error type events}
    \item \texttt{Events on alerting entities or their dependencies}
    \item \texttt{Pod lifecycle failures (CrashLoopBackOff, OOMKilled, Evicted)}
    \item \texttt{Scheduling failures (FailedScheduling, FailedMount)}
    \item \texttt{Probe failures (Unhealthy, ProbeWarning)}
    \item \texttt{Resource issues (FailedCreate, QuotaExceeded)}
    \item \texttt{Deployments/restarts preceding alerts}
\end{itemize}

\vspace{0.3em}
\texttt{\textbf{EXCLUDE:} Normal successful operations (Scheduled, Started, Created) unless systematic churn or temporally preceding alerts.}

\vspace{0.3em}
\texttt{\textbf{Prioritization:} Return at most \{\{max\_per\_page\}\} events. Rank by: (1) events preceding alerts, (2) events on alerting entities, (3) failure events, (4) config changes. Prefer semantic diversity---avoid redundant events conveying the same information.}

\vspace{0.3em}
\texttt{\textbf{Output:} JSON with relevant\_events array, total\_relevant\_before\_limit, reasoning.}
\end{promptbox}

\subsubsection{Spec Change Filtering}

Configuration changes are among the most common root causes. We filter spec changes with similar pagination.

\begin{promptbox}{filter\_spec\_changes}
\scriptsize
\texttt{You are filtering Kubernetes spec changes to identify those relevant to an ongoing incident investigation. Configuration changes are one of the biggest failure causes.}

\vspace{0.3em}
\texttt{\textbf{Prioritization Guidance:}}
\begin{enumerate}[nosep,leftmargin=1.5em]
    \item \texttt{\textbf{Highest:} Changes just before alerts (within minutes)---prime root cause suspects}
    \item \texttt{\textbf{Medium:} Changes during incident window}
    \item \texttt{\textbf{Lower:} Changes well before/after incident}
\end{enumerate}

\vspace{0.3em}
\texttt{Key Question: ``Did this change happen before the symptoms started?'' If yes, strong candidate for root cause.}

\vspace{0.3em}
\texttt{\textbf{Output:} JSON with relevant\_changes array (max \{\{max\_per\_page\}\}), total\_relevant\_before\_limit, reasoning.}
\end{promptbox}

\subsubsection{Candidate Selection (Bootstrap)}

With filtered events and spec changes, the bootstrap prompt selects initial investigation candidates.

\begin{promptbox}{bootstrap}
\scriptsize
\texttt{You are an SRE agent. Review active alerts and select initial candidate entities for investigation.}

\vspace{0.3em}
\texttt{\textbf{Strategy (Golden Signal Priority):}}
\begin{enumerate}[nosep,leftmargin=1.5em]
    \item \texttt{Identify minimum set of alerts that cannot be explained by other alerts/events}
    \item \texttt{Ignore positive goodput/watchdog alerts}
    \item \texttt{Infrastructure alerts (OOM, throttling) help explain golden signals}
\end{enumerate}

\vspace{0.3em}
\texttt{\textbf{Strict Ordering by Causal Evidence Strength:}}
\begin{itemize}[nosep,leftmargin=1em]
    \item \texttt{\textbf{STRONGEST (FIRST):} Entity with spec change preceding alert start}
    \item \texttt{\textbf{STRONG:} K8s event (OOMKilled, CrashLoopBackOff) on alerting entity}
    \item \texttt{\textbf{MEDIUM:} Change on entity with known dependency relationship}
    \item \texttt{\textbf{WEAKEST:} Alerting service with NO associated change---likely symptom}
\end{itemize}

\vspace{0.3em}
\texttt{\textbf{Causal Path Requirement:} A spec change is only relevant if there's a causal path to the alerting application (dependency, traffic source, config, fault injection).}

\vspace{0.3em}
\texttt{\textbf{Output:} JSON with initial\_candidates array (node\_id, start\_time, end\_time, alert\_type, reason\_for\_selection, identified\_facts, identified\_events), reasoning.}
\end{promptbox}

\subsection{Abductive Policy Prompt}
\label{appendix:explore_prompt}

The explore prompt implements local inference, reasoning over both local evidence and neighbor messages.

\begin{promptbox}{explore (Abductive Policy $\pi_{\text{abd}}$)}
\scriptsize
\texttt{You are an SRE support engineer doing RCA using the EoG algorithm.}

\vspace{0.3em}
\texttt{\textbf{Objective:} Help the controller converge to a small root-cause frontier. Do NOT enumerate all impacted entities.}

\vspace{0.3em}
\texttt{\textbf{Definitions:}}
\begin{itemize}[nosep,leftmargin=1em]
    \item \texttt{\textbf{Healthy (L=0):} Operating normally in the alert window}
    \item \texttt{\textbf{Symptom (L=0):} Degraded, but evidence points to upstream cause}
    \item \texttt{\textbf{Origin (L=1):} Has citable change (config, deployment, resource) that temporally precedes and explains incident}
    \item \texttt{\textbf{Defer (L=2):} Evidence inconclusive}
\end{itemize}

\vspace{0.3em}
\texttt{\textbf{Messages from Neighbors (Inbox):} When neighbor claims SYMPTOM, consider if YOUR component caused their failure. When neighbor claims ORIGIN, consider if YOUR issues are symptoms of their fault.}

\vspace{0.3em}
\texttt{\textbf{Core Investigation Principles:}}
\begin{enumerate}[nosep,leftmargin=1.5em]
    \item \texttt{\textbf{``What Changed'' Discipline:} A node cannot be Origin if nothing changed within it}
    \item \texttt{\textbf{Transitive Traffic Amplification:} Resource exhaustion without local change $\to$ suspect traffic surge}
    \item \texttt{\textbf{Differential Observability:} Compare replicas, time windows}
    \item \texttt{\textbf{Occam's Razor + Multiple Causes:} Prefer unified explanation, accept independent causes when evidence supports}
\end{enumerate}

\vspace{0.3em}
\texttt{\textbf{Hard Rules:}}
\begin{itemize}[nosep,leftmargin=1em]
    \item \texttt{Evidence required for every decision (cite logs/events/metrics/traces)}
    \item \texttt{Never revisit entities in Visited set}
    \item \texttt{Max 2 next\_candidates per step}
    \item \texttt{One direction at a time: upstream $|$ downstream $|$ ownership $|$ infrastructure}
\end{itemize}

\vspace{0.3em}
\texttt{\textbf{Output:} JSON with direction, local\_origin (Healthy$|$Symptom$|$Origin$|$Defer), reasoning, propagations array, next\_candidates array.}
\end{promptbox}

\subsection{Finalize Prompt}
\label{appendix:finalize_prompt}

When the event queue empties or budget exhausts, the finalize prompt synthesizes findings.

\begin{promptbox}{finalize}
\scriptsize
\texttt{Summarize the investigation, identifying Root Causes (Frontier) and Causal Graph ($G_S$). If no frontier found, make the most educated guess given all evidence.}

\vspace{0.3em}
\texttt{\textbf{Critical Distinction:}}
\begin{itemize}[nosep,leftmargin=1em]
    \item \texttt{\textbf{Root Cause} (contributing\_factor: true) = Entity that CAUSED the problem: config changes, deployments, infrastructure failures, fault injections, traffic pattern changes}
    \item \texttt{\textbf{Symptom} (contributing\_factor: false) = Entity that EXHIBITED the problem: elevated error rates, failures due to upstream causes}
\end{itemize}

\vspace{0.3em}
\texttt{\textbf{Validation Checkpoint:} For each contributing\_factor: true entity, answer: (1) What specifically changed? (2) When did it change? (3) How does this change explain the symptoms? If you cannot answer all three, it's likely a symptom.}

\vspace{0.3em}
\texttt{\textbf{Contributing Factor Rules:}}
\begin{itemize}[nosep,leftmargin=1em]
    \item \texttt{\textbf{True (Irreducible):} Origin in Frontier with no incoming causal edge from another Origin}
    \item \texttt{\textbf{False (Derived):} Symptom affected by propagation; failure explained by upstream}
\end{itemize}

\vspace{0.3em}
\texttt{\textbf{Output:} JSON with entities array (name, contributing\_factor, reasoning, evidence), propagations array (source, target, condition, effect), alerts\_explained array.}
\end{promptbox}

\subsection{Implementation Notes}
\label{appendix:impl_notes}

The system comprises approximately 3,500 lines of Rust with prompts maintained as Handlebars templates. Key implementation choices:

\begin{itemize}[nosep]
    \item \textbf{Temperature Zero:} All LLM calls use temperature 0 for determinism.
    \item \textbf{Structured Output:} All prompts request JSON validated against schemas. Invalid responses trigger a single retry with error correction.
    \item \textbf{Token Budgeting:} The Context Contract implements token-aware pagination. When filtered events/changes exceed budgets, tournament-style reduction selects the most relevant items.
    \item \textbf{Entity Normalization:} All prompts mandate \texttt{namespace/Kind/name} format for consistent deduplication and graph construction.
\end{itemize}

The implementation is available as part of the open-source Codex project~\cite{openai_codex}.

\subsection{Properties and Correctness}
\label{appendix:eog_properties}

\subsubsection{Minimality and Correctness}

The algorithm seeks a \emph{Frontier} $\mathcal{F}$ of irreducible origins.
\begin{equation}
\begin{aligned}
\mathcal{F} = \{v \in V_S :\;& L_v = \textsc{Origin} \land \nexists u \in V_S \\
& \text{s.t. } L_u = \textsc{Origin} \land u \rightsquigarrow_{E_S} v\}
\end{aligned}
\end{equation}
By definition, $\mathcal{F}$ is minimal relative to the discovered graph $G_S$. If no node is labeled \textsc{Origin} (i.e., $\mathcal{F} = \emptyset$), the system falls back to best-effort ranking of candidates based on accumulated evidence.
The correctness relies on three assumptions:
\begin{enumerate}[nosep]
    \item \textbf{Completeness}: Relevant data exists and is retrievable via CxC.
    \item \textbf{Soundness}: The policy $\pi_{\text{abd}}$ correctly infers local dependencies.
    \item \textbf{Connectivity}: Origins are reachable from anomalous entities $V_\mathcal{O}$ via the topology or discovered edges.
\end{enumerate}
If these hold, the message passing ensures that evidence from origins drives the observed symptoms, correcting any initial misclassifications caused by the exploration order.

\paragraph{On the role of the deterministic controller.}
The Soundness assumption merits clarification: EoG does not eliminate LLM stochasticity---belief assignments remain LLM-dependent. The deterministic controller ensures systematic exploration, belief bookkeeping, and propagation logic, but does not guarantee that local inferences are correct. The architecture's value lies in \emph{bounding} the scope of each LLM decision to local context and \emph{enabling revision} when contradictory evidence emerges via message passing. Reliability gains (Majority@$k$ improvements) arise from this structured exploration and revision mechanism, not from eliminating stochasticity in individual inferences.

\subsubsection{Termination}

Termination is guaranteed by the damping heuristic. If a node oscillates labels more than $k_{\text{thresh}}$ times, it is forced to \textsc{Defer} (an absorbing state). This bounds the total number of inference steps to $O(k_{\text{thresh}} \cdot |V_S|)$. Cycles in the topology are allowed, but the explanatory graph $E_S$ is directed by evidence. When evidence is conflicting, oscillation can occur; damping converts this to the absorbing \textsc{Defer} state and guarantees a fixed point when no messages remain.

\subsubsection{Handling Inconclusive Results}

The agent may not always produce a definitive diagnosis. Two failure modes are possible: (1)~\emph{evidence absence}---the relevant observability data is missing; and (2)~\emph{evidence blindness}---the policy $\pi_{\text{abd}}$ lacks the domain knowledge to recognize available evidence as diagnostically relevant. In either case, nodes are labeled \textsc{Defer}. If no \textsc{Origin} is identified at termination (i.e., frontier $\mathcal{F} = \emptyset$), the Controller falls back to a best-effort ranking of candidate nodes based on accumulated evidence, explicitly signaling uncertainty to the operator.

\section{Analyzing Controller Failures}
\label{appendix:controller_analysis}

The ReAct paradigm conflates \emph{reasoning} (what to investigate) with \emph{execution} (how to invoke tools).
This overloading of the LLM leads to measurable controller instability.
We analyze three failure modes---plan abandonment, tool repetition, and syntactic failures---that motivate the separation of concerns in EoG.

\begin{figure*}[t]
  \centering
  \begin{subfigure}{0.32\textwidth}
    \centering
    \includegraphics[width=\linewidth]{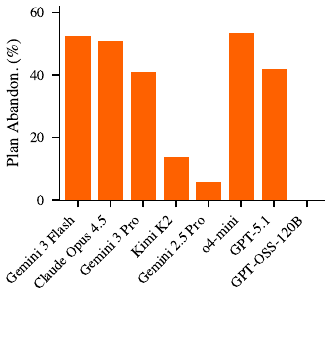}
    \caption{Plan abandonment rate.}
    \label{fig:plan_abandonment}
  \end{subfigure}
  \hfill
  \begin{subfigure}{0.32\textwidth}
    \centering
    \includegraphics[width=\linewidth]{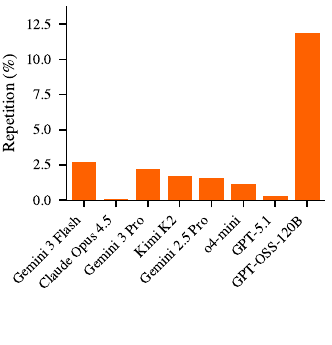}
    \caption{Tool repetition rate.}
    \label{fig:tool_repetition}
  \end{subfigure}
  \hfill
  \begin{subfigure}{0.32\textwidth}
    \centering
    \includegraphics[width=\linewidth]{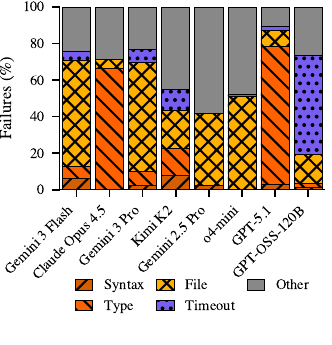}
    \caption{Tool failure categories.}
    \label{fig:tool_failures}
  \end{subfigure}
  \caption{Controller brittleness in ReAct agents. (a)~Plan abandonment rate: frequency with which agents discard stated investigation plans. (b)~Tool repetition rate: percentage of tool calls that duplicate previous invocations. (c)~Tool failure categories: distribution of syntactic vs.\ semantic errors.}
  \label{fig:controller_brittleness}
\end{figure*}

\paragraph{Plan abandonment.}
Figure~\ref{fig:plan_abandonment} quantifies how often agents abandon their stated investigation plans.
o4-mini abandons plans in over 50\% of cases; even stronger models like Claude Opus 4.5 exhibit abandonment rates over 50\%.
Plan abandonment occurs for three reasons:
(1)~agents fail to revise plans as new evidence emerges, leaving stale steps that no longer apply~\cite{gandhi2025agentsastraycoursecorrectingswe};
(2)~agents forget earlier plan steps as context grows (context rot)~\cite{liu2024lost};
and (3)~agents prematurely conclude they have found the cause and skip remaining steps, converging to a minimal explanation before sufficient evidence is gathered.
This instability undermines auditability: the agent's stated reasoning diverges from its actual behavior.

\paragraph{Tool repetition.}
Agents exhibit degenerate loops where they repeatedly invoke the same tool calls.
Figure~\ref{fig:tool_repetition} shows that GPT-OSS-120B has a repetition rate of about 12\%, consuming context budget without advancing the investigation.
Tool repetition can occur for two reasons:
(1)~re-executing a failed tool call without correcting the arguments or function signature;
or (2)~re-fetching data to analyze it differently in light of new evidence.
The latter is potentially useful---revisiting data with updated hypotheses is a valid diagnostic strategy---but we find no evidence of productive re-analysis in our trajectories; the vast majority of repetitions are failed-execution loops.
This redundancy compounds with trajectory length, leaving less capacity for novel exploration.

\paragraph{Syntactic failures.}
Figure~\ref{fig:tool_failures} categorizes tool invocation failures.
A substantial fraction are strictly syntactic---malformed arguments, type errors, invalid parameter combinations---rather than semantic reasoning failures~\cite{chen2024tevalevaluatingtoolutilization}.
These errors indicate that LLMs are suboptimal drivers for precise tool orchestration when simultaneously reasoning about complex domains.
In contrast, conventional non-deterministic programs tend to excel in tool invocation and execution when they have access to the right contextual information.

\subsection{MAST Failure Mode Taxonomy}
\label{appendix:mast_taxonomy}

We use MAST (Multi-Agent System Trajectories)~\cite{mast2025} to systematically categorize controller failures in our agent trajectories. MAST provides a structured taxonomy that classifies failures across three inter-agent conversation stages: Pre-Execution, Execution, and Post-Execution.

\begin{figure*}[t]
    \centering
    \includegraphics[width=0.75\linewidth]{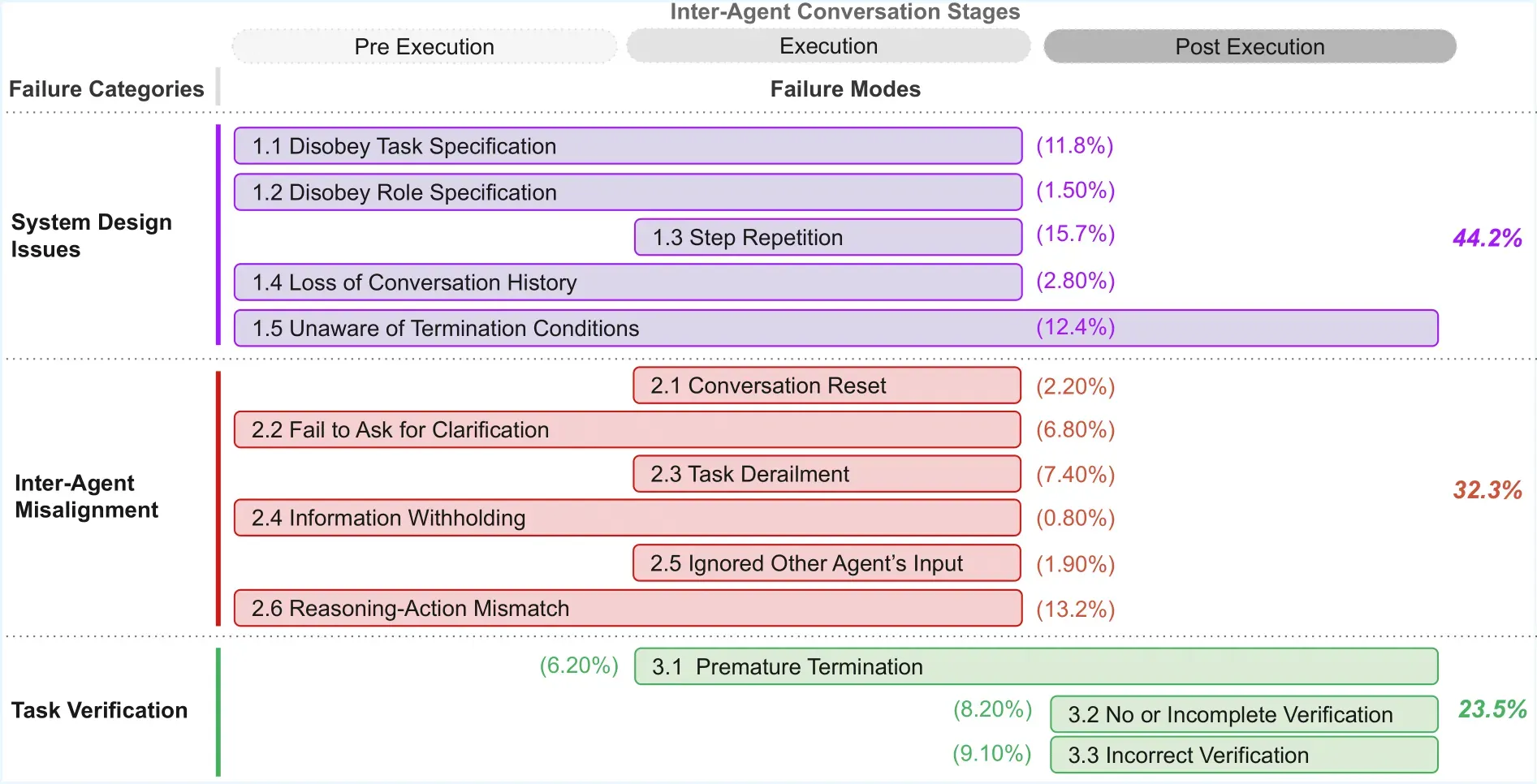}
    \caption{MAST failure mode taxonomy. Failures are categorized into three main categories: \textbf{System Design Issues} (44.2\%)---failures stemming from how the agent system is architected; \textbf{Inter-Agent Misalignment} (32.3\%)---failures in communication and coordination; and \textbf{Task Verification} (23.5\%)---failures in validating task completion.}
    \label{fig:mast_taxonomy}
\end{figure*}

Figure~\ref{fig:mast_taxonomy} shows the taxonomy with prevalence rates across our trajectories:

\paragraph{System Design Issues (44.2\%).}
The largest category encompasses architectural failures:
\begin{itemize}[nosep]
    \item \textbf{1.1 Disobey Task Specification (11.8\%)}: Agent ignores explicit task instructions.
    \item \textbf{1.2 Disobey Role Specification (1.5\%)}: Agent acts outside its designated role.
    \item \textbf{1.3 Step Repetition (15.7\%)}: Agent repeats previously executed steps without progress.
    \item \textbf{1.4 Loss of Conversation History (2.8\%)}: Agent loses track of prior context.
    \item \textbf{1.5 Unaware of Termination Conditions (12.4\%)}: Agent fails to recognize when task is complete.
\end{itemize}

\paragraph{Inter-Agent Misalignment (32.3\%).}
Communication and coordination failures:
\begin{itemize}[nosep]
    \item \textbf{2.1 Conversation Reset (2.2\%)}: Agent restarts conversation inappropriately.
    \item \textbf{2.2 Fail to Ask for Clarification (6.8\%)}: Agent proceeds despite ambiguity.
    \item \textbf{2.3 Task Derailment (7.4\%)}: Agent deviates from the assigned task.
    \item \textbf{2.4 Information Withholding (0.8\%)}: Agent fails to share relevant information.
    \item \textbf{2.5 Ignored Other Agent's Input (1.9\%)}: Agent disregards feedback.
    \item \textbf{2.6 Reasoning-Action Mismatch (13.2\%)}: Agent's stated reasoning contradicts its actions.
\end{itemize}

\paragraph{Task Verification (23.5\%).}
Failures in validating task completion:
\begin{itemize}[nosep]
    \item \textbf{3.1 Premature Termination (6.2\%)}: Agent stops before task completion.
    \item \textbf{3.2 No or Incomplete Verification (8.2\%)}: Agent fails to verify its output.
    \item \textbf{3.3 Incorrect Verification (9.1\%)}: Agent incorrectly validates erroneous output.
\end{itemize}

\begin{figure*}[t]
    \centering
    \includegraphics[width=\linewidth]{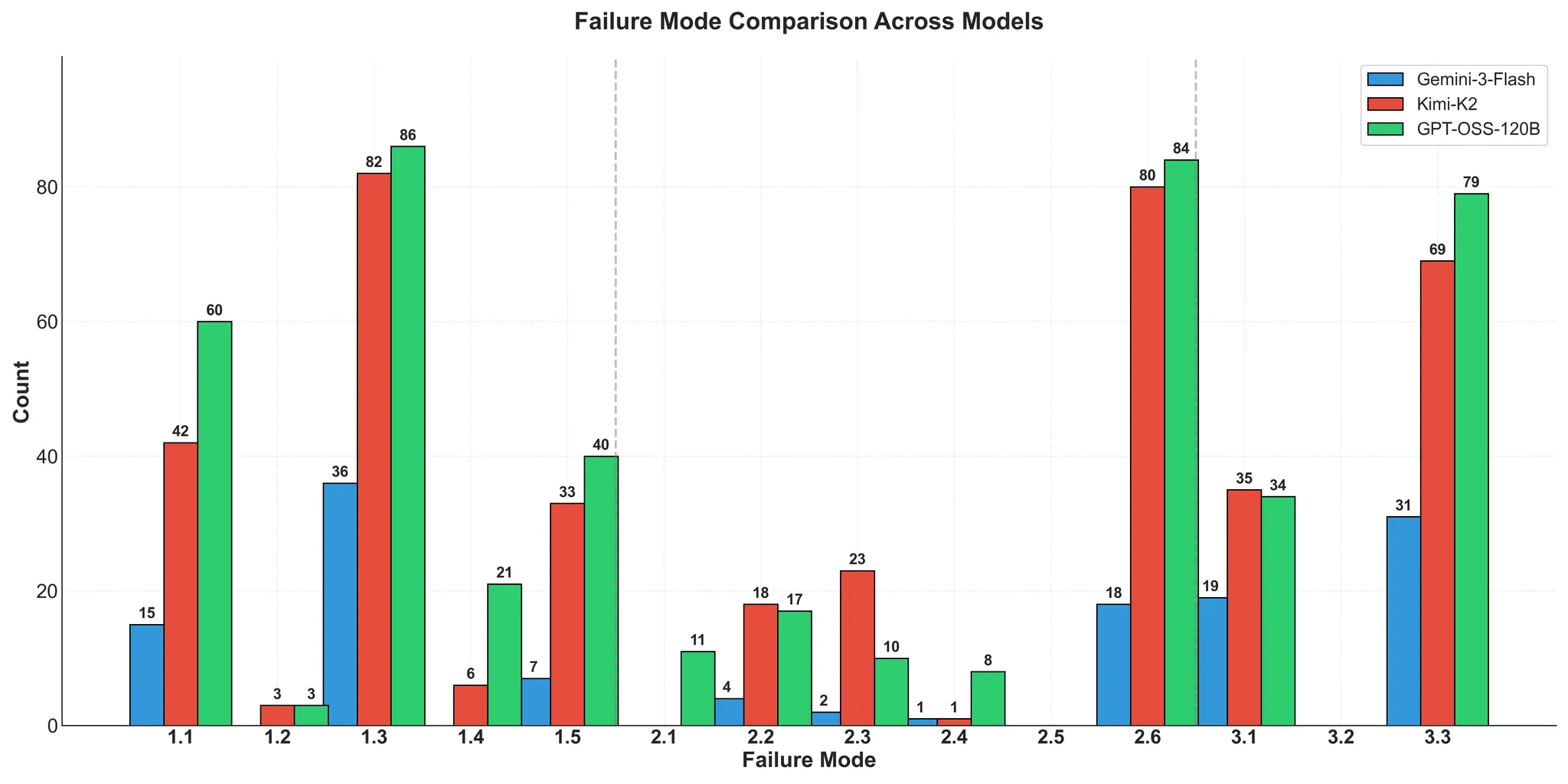}
    \caption{Failure mode counts across models. GPT-OSS-120B and Kimi-K2 exhibit higher failure counts across most categories compared to Gemini-3-Flash, with Step Repetition (1.3) and Reasoning-Action Mismatch (2.6) being the most prevalent failure modes.}
    \label{fig:mast_comparison}
\end{figure*}

Figure~\ref{fig:mast_comparison} compares failure mode prevalence across three models. Key observations:
(1)~Step Repetition (1.3) is the most common failure across all models, with GPT-OSS-120B showing 86 instances versus 36 for Gemini-3-Flash.
(2)~Reasoning-Action Mismatch (2.6) shows similar patterns, indicating a fundamental limitation of the ReAct paradigm where LLMs must simultaneously plan and execute.
(3)~Gemini-3-Flash exhibits consistently lower failure counts, correlating with its higher diagnostic accuracy.

\begin{figure*}[t]
    \centering
    \includegraphics[width=0.75\linewidth]{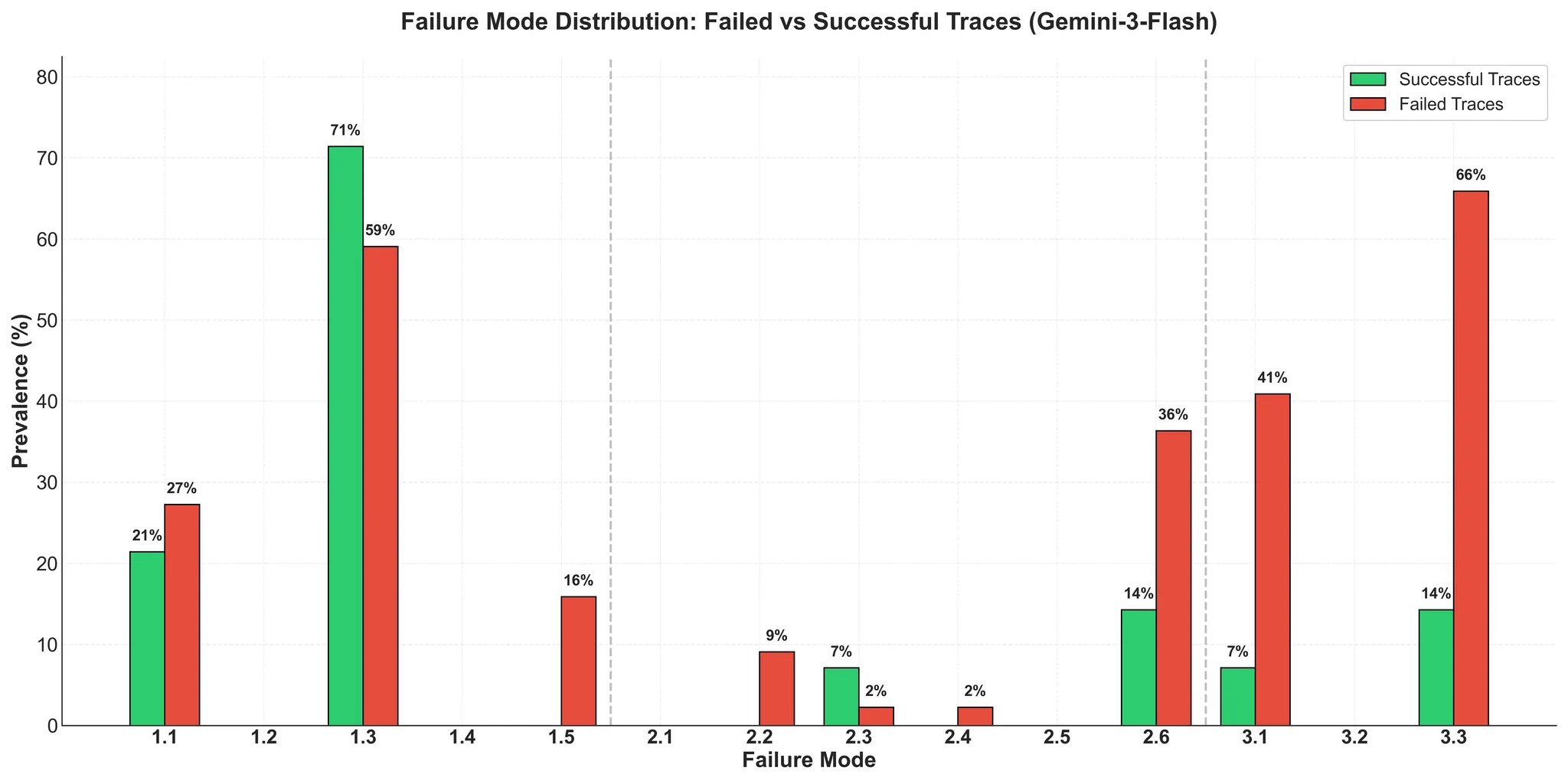}
    \caption{Failure mode distribution in successful vs.\ failed traces for Gemini-3-Flash. Green bars show prevalence in successful traces; red bars show prevalence in failed traces.}
    \label{fig:mast_gemini}
\end{figure*}

\begin{figure*}[t]
    \centering
    \includegraphics[width=0.75\linewidth]{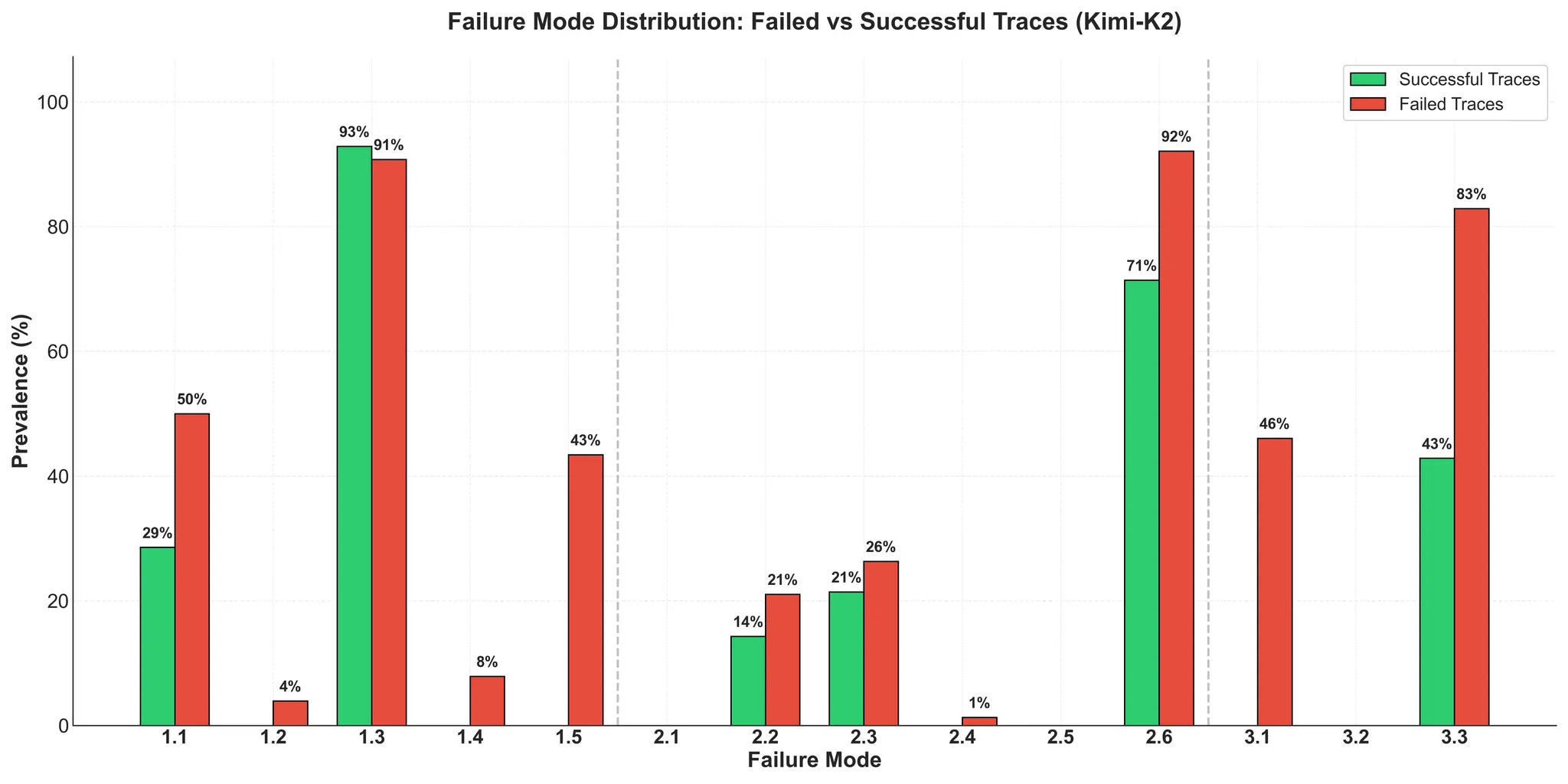}
    \caption{Failure mode distribution in successful vs.\ failed traces for Kimi-K2. Green bars show prevalence in successful traces; red bars show prevalence in failed traces.}
    \label{fig:mast_kimi}
\end{figure*}

\begin{figure*}[t]
    \centering
    \includegraphics[width=0.75\linewidth]{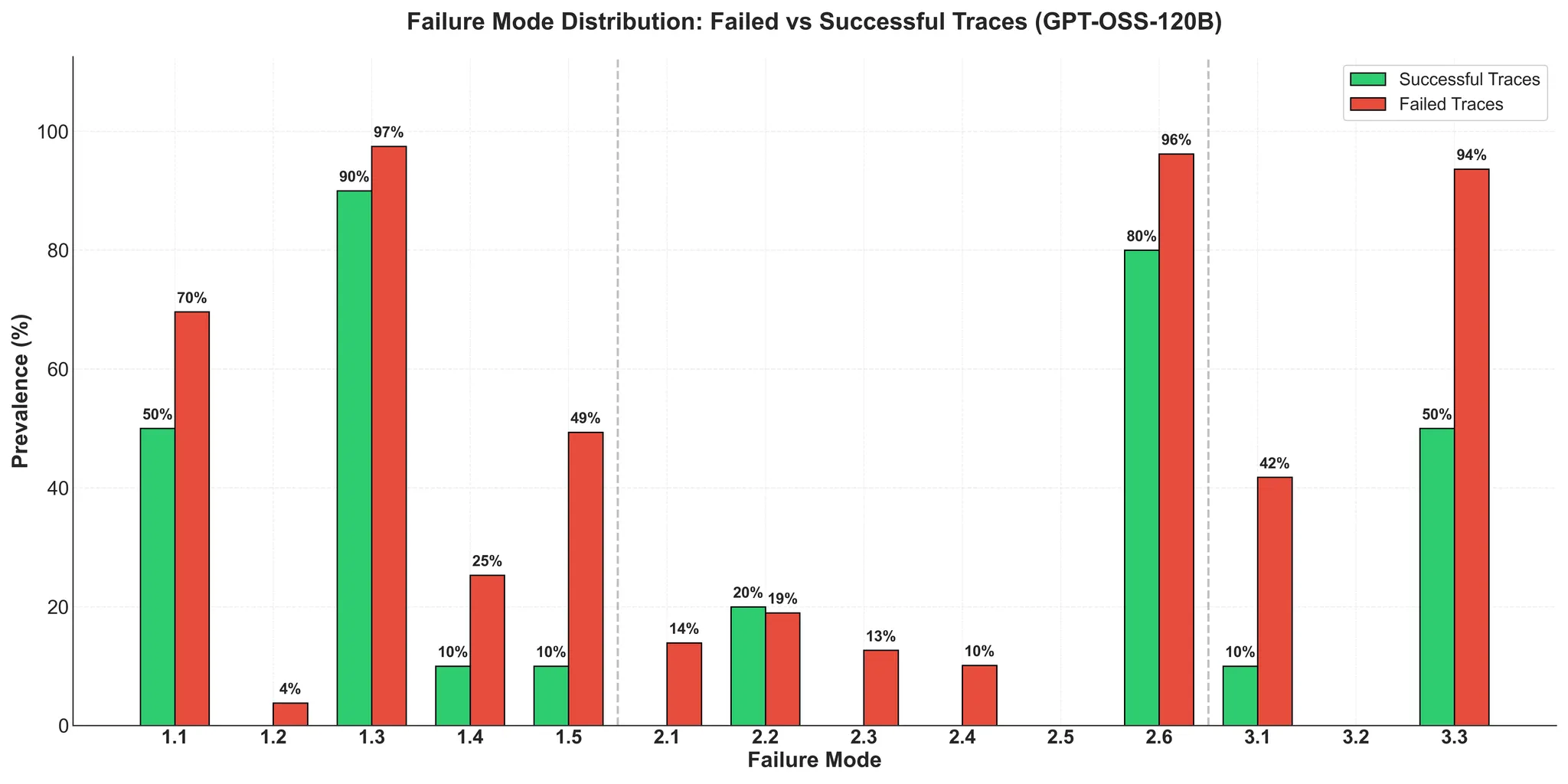}
    \caption{Failure mode distribution in successful vs.\ failed traces for GPT-OSS-120B. Green bars show prevalence in successful traces; red bars show prevalence in failed traces.}
    \label{fig:mast_gpt}
\end{figure*}

Figures~\ref{fig:mast_gemini}--\ref{fig:mast_gpt} reveal which failure modes differentiate successful from failed traces:
\begin{itemize}[nosep]
    \item \textbf{Discriminative failures}: Incorrect Verification (3.3) shows the largest gap between failed and successful traces across all models (66\% vs 14\% for Gemini-3-Flash, 83\% vs 43\% for Kimi-K2, 94\% vs 50\% for GPT-OSS-120B). This indicates that verification failures are strongly predictive of task failure.
    \item \textbf{Non-discriminative failures}: Step Repetition (1.3) appears at high rates in both successful and failed traces (71\% vs 59\% for Gemini-3-Flash), suggesting that some repetition may be a natural part of iterative exploration rather than a direct cause of failure.
    \item \textbf{Model-specific patterns}: GPT-OSS-120B shows elevated failure rates across nearly all categories in failed traces, while Gemini-3-Flash maintains lower baseline rates even in failed cases.
\end{itemize}

These findings motivate the core architectural decision in EoG: by separating deterministic control (state management, termination detection, verification) from non-deterministic reasoning (causal inference), we eliminate the System Design Issues (44.2\%) and Task Verification failures (23.5\%) that account for over two-thirds of observed failures.

\section{LLM-as-a-Judge Evaluation Framework}
\label{appendix:evaluation}

We employ an LLM-as-a-Judge (LAAJ) evaluation framework to assess agent outputs against ground truth root cause analysis. The judge evaluates two primary metrics---\textbf{Root Cause Entity Identification} (Precision, Recall, F1) and \textbf{Root Cause Reasoning Accuracy}---in a single LLM call.

\paragraph{Evaluation cost considerations.}
Each agent run produces a structured diagnosis that must be evaluated against ground truth with semantic matching (entity aliasing, reasoning quality assessment).
The LAAJ evaluation requires an LLM call per (scenario, run) pair, incurring non-trivial cost at scale.
For our benchmark of 50 scenarios with $k=5$ runs per model, evaluating a single model requires 250 LAAJ calls.
Due to this cost, we focus our detailed analysis on a representative subset of frontier models (Gemini 2.5 Pro/Flash, Claude Opus 4.5, o4-mini) that span the capability spectrum, while reporting aggregate metrics for additional models.

\subsection{Entity Mapping Challenge}
\label{appendix:entity_mapping_example}

To understand why entity normalization and mapping is critical, consider Scenario-1 from our benchmark involving a traffic overload incident.

\subsubsection{Ground Truth Structure}

The ground truth defines entities using semantic identifiers with regex-based filters:

\begin{lstlisting}[basicstyle=\scriptsize\ttfamily,frame=single,caption={Ground Truth Entity Groups (Scenario-1)}]
groups:
  - id: load-generator-pod-1
    kind: Pod
    filter: [load-generator-.*]
    namespace: otel-demo
    root_cause: true
  - id: load-generator-deployment-1
    kind: Deployment
    filter: [load-generator\b]
    namespace: otel-demo
    root_cause: true
  - id: flagd-config-1
    kind: ConfigMap
    filter: [flagd-config\b]
    namespace: otel-demo
    root_cause: true
  - id: frontend-proxy-service-1
    kind: Service
    filter: [frontend-proxy\b]
    namespace: otel-demo
\end{lstlisting}

\subsubsection{Alias Groups}

Multiple Kubernetes objects represent the same logical service. The ground truth captures this through alias groups:

\begin{lstlisting}[basicstyle=\scriptsize\ttfamily,frame=single,caption={Ground Truth Aliases}]
aliases:
  - [load-generator-service-1, load-generator-pod-1, 
     load-generator-deployment-1]
  - [frontend-proxy-service-1, frontend-proxy-pod-1]
\end{lstlisting}

If a model identifies \texttt{otel-demo/Service/load-generator} as a root cause, it receives credit because it is an alias of \texttt{load-generator-pod-1}.

\subsubsection{Propagation Chain}

The ground truth defines the causal chain showing how the fault propagated:

\begin{lstlisting}[basicstyle=\scriptsize\ttfamily,frame=single,caption={Ground Truth Propagation Chain}]
propagations:
  - source: flagd-config-1
    target: load-generator-deployment-1
    condition: flagd-config updated to enable higher traffic
    effect: load-generator deployment applies new config
  - source: load-generator-deployment-1
    target: load-generator-pod-1
    condition: updated deployment rolls out pods
    effect: load-generator pod configured with higher users
  - source: load-generator-pod-1
    target: load-generator-service-1
    condition: pod generates high number of requests
    effect: service receives high request volume
  - source: load-generator-service-1
    target: frontend-proxy-service-1
    condition: frontend-proxy overloaded
    effect: Error rate and latency above threshold
\end{lstlisting}

\subsubsection{The Mapping Challenge}

Actual Kubernetes object names include generated suffixes. The evaluation must map agent-predicted entities to ground truth semantic IDs using regex filters. For example:
\begin{itemize}[nosep]
    \item \texttt{load-generator-pod-1} matches \texttt{otel-demo/Pod/load-generator-c7cbc4c99-xyz12}
    \item \texttt{flagd-config-1} matches \texttt{otel-demo/ConfigMap/flagd-config}
    \item \texttt{load-generator-deployment-1} matches \texttt{otel-demo/Deployment/load-generator}
\end{itemize}

\subsection{Judge System Prompt}
\label{appendix:judge_system}

The LAAJ system prompt defines the evaluation protocol:

\begin{promptbox}{LAAJ System Prompt}
\scriptsize
\texttt{You are an expert AI evaluator specializing in Root Cause Analysis (RCA) for complex software systems. Your role is to act as a fair, consistent, and precise judge when evaluating RCA model outputs.}

\vspace{0.2em}
\texttt{You will be provided with: (1) A \textbf{Ground Truth (GT)} JSON object: true root cause(s), failure propagation path(s), and entity definitions. (2) A \textbf{Generated Response} JSON object: output from an RCA model to evaluate.}

\vspace{0.2em}
\texttt{\textbf{Phase 1: Normalization and Entity Mapping} --- Before any scoring, normalize and map entities from Generated Response to Ground Truth based on \textbf{explicit evidence} from entity metadata. Do not infer mappings based on position in propagation chain.}

\vspace{0.2em}
\texttt{\textbf{Definition of a Confident Match:} A Generated Response entity is a confident match ONLY IF its \texttt{name} field clearly corresponds to the \texttt{filter} and \texttt{kind} of the GT entity.}

\vspace{0.2em}
\texttt{\textbf{Entity Name Format:} Generated Response entities use format: \texttt{namespace/Kind/name}. Examples: \texttt{otel-demo/Deployment/flagd}, \texttt{otel-demo/Service/frontend}.}

\vspace{0.2em}
\texttt{\textbf{CRITICAL - Alias Handling:} If a predicted entity matches an ALIAS of a root cause entity, it MUST be counted as correct. The \texttt{GT.aliases} field contains arrays of equivalent entity IDs. If ANY entity in an alias group has \texttt{root\_cause: true}, then matching ANY entity in that alias group counts as correctly identifying the root cause.}

\vspace{0.2em}
\texttt{\textbf{Example:} Model entity \texttt{otel-demo/Service/adservice} matches GT entity \texttt{ad-service-1} with filter \texttt{[".*adservice\textbackslash\textbackslash b"]}.}
\end{promptbox}

\subsection{Evaluation Metrics}
\label{appendix:metrics}

The LAAJ framework evaluates two primary metrics:
\begin{itemize}[nosep]
    \item \textbf{Root Cause Entity} (Precision, Recall, F1): Match \texttt{contributing\_factor=true} entities against GT root causes with alias support.
    \item \textbf{Root Cause Reasoning} (Average): Semantic similarity of condition descriptions using 0/0.5/1 scoring.
\end{itemize}

\subsubsection{Root Cause Entity Identification}

\begin{metricbox}{Root Cause Entity Scoring}
\small
\texttt{\textbf{Method:}}
\begin{enumerate}[nosep,leftmargin=1.5em]
    \item \texttt{Extract all entities with \texttt{root\_cause: true} in GT.groups.}
    \item \texttt{\textbf{Handle Aliases:} Check GT.aliases for any alias groups containing a root cause entity. ALL entities in such groups are valid root cause matches.}
    \item \texttt{Extract all entities with \texttt{contributing\_factor: true} from Generated Response, preserving order.}
    \item \texttt{Normalize entities to compare against GT root causes.}
    \item \texttt{For EACH predicted entity (in order), determine if it matches any GT root cause entity OR any alias.}
    \item \texttt{Output per-entity match list showing which predicted entities matched GT entities.}
    \item \texttt{Calculate Recall = Correct / Total GT (count each alias group only once).}
    \item \texttt{Calculate Precision = Correct / Total Predicted.}
    \item \texttt{Calculate F1 = (2 $\times$ Recall $\times$ Precision) / (Recall + Precision).}
\end{enumerate}

\texttt{\textbf{Output:} List of GT entities, ordered list of predicted entities with per-entity match boolean, Precision, Recall, and F1.}
\end{metricbox}

\subsubsection{Root Cause Reasoning Accuracy}

\begin{metricbox}{Root Cause Reasoning Scoring}
\small
\texttt{\textbf{Method:}}
\begin{enumerate}[nosep,leftmargin=1.5em]
    \item \texttt{Extract the \texttt{condition} of every correct root cause entity from Generated Response.}
    \item \texttt{Extract the \texttt{condition} of corresponding entities from GT.}
    \item \texttt{For each correct root cause entity, compare conditions:}
    \begin{itemize}[nosep,leftmargin=1em]
        \item \texttt{Score 1.0: Model's reason is semantically equivalent to GT's condition}
        \item \texttt{Score 0.5: Model's condition describes similar resource/action but at least one key detail is missing or imprecise (e.g., GT=``configmap Y featureflag X set'', Model=``configmap updated'')}
        \item \texttt{Score 0.0: Incorrect or absent model condition}
    \end{itemize}
    \item \texttt{Calculate average of all entity reasoning scores.}
    \item \texttt{If no root cause entity correctly identified, score = 0.}
\end{enumerate}

\texttt{\textbf{Output:} Average reasoning score.}
\end{metricbox}

\subsection{Evaluation Prompt Template}

\begin{promptbox}{Evaluation User Prompt}
\small
\texttt{Given the following Ground Truth (GT) and Generated Response, evaluate the response according to the scoring rubric.}

\vspace{0.3em}
\texttt{\#\# Ground Truth (GT):}\\
\texttt{\{ground\_truth\}}

\vspace{0.3em}
\texttt{\#\# Generated Response:}\\
\texttt{\{generated\_response\}}

\vspace{0.3em}
\texttt{\#\# Task:}
\begin{enumerate}[nosep,leftmargin=1.5em]
    \item \texttt{Perform Phase 1: Normalize and map entities.}
    \item \texttt{Calculate ROOT\_CAUSE\_ENTITY metrics (Precision, Recall, F1).}
    \item \texttt{Calculate ROOT\_CAUSE\_REASONING score.}
    \item \texttt{\textbf{IMPORTANT}: Use \texttt{calculator\_tool} for ALL mathematical calculations.}
\end{enumerate}

\texttt{\textbf{CRITICAL:} Final response MUST be ONLY the JSON object with scores.}
\end{promptbox}

\end{document}